\documentclass[final,3p,times]{elsarticle}

\usepackage{xcolor}
\usepackage{url}
\usepackage{booktabs} 

\usepackage{amssymb}
\usepackage{amsthm}
\usepackage{amsmath}
\newtheorem{theorem}{Theorem}[section]
\newtheorem{lemma}[theorem]{Lemma}

\newtheorem{corollary}[theorem]{Corollary}
\newtheorem{assumption}{Assumption}
\newtheorem{definition}{Definition}

\makeatletter
\renewenvironment{proof}[1][\proofname]{\par
  \pushQED{\qed}%
  \normalfont \topsep6\p@\@plus6\p@\relax
  \trivlist
  \item[\hskip\labelsep
        \bfseries 
    #1]\ignorespaces 
}{%
  \popQED\endtrivlist\@endpefalse
}
\makeatother

\usepackage{enumitem}

\usepackage[linesnumbered,ruled,vlined]{algorithm2e}
\SetKwBlock{Loop}{Loop}{}

\journal{Nonlinear Analysis: Hybrid Systems}

\begin{document}

\begin{frontmatter}



\title{Accelerating Hybrid Model Predictive Control using Warm-Started \\ Generalized Benders Decomposition}




\author[1,2,3]{Xuan Lin}

\affiliation[1]{organization={Independent Researcher}, country={USA}}

\affiliation[2]{organization={Department of Mechanical Engineering, University of California, Los Angeles},
    city={Los Angeles},
    state={CA},
    postcode={90095},
    country={USA}}

\affiliation[3]{organization={School of Mechanical Engineering, Georgia Institute of Technology},
    city={Atlanta},
    state={GA},
    postcode={30318},
    country={USA}}
    
\begin{abstract}
Hybrid model predictive control with both continuous and discrete variables is widely applicable to robotic control tasks, especially those involving contacts with the environment. Due to combinatorial complexity, the solving speed of hybrid MPC can be insufficient for real-time applications. In this paper, we propose a hybrid MPC algorithm based on Generalized Benders Decomposition. The algorithm enumerates and stores cutting planes online inside a finite buffer and transfers them across MPC iterations to provide warm-starts for new problem instances, significantly enhancing solving speed. We theoretically analyze this warm-starting performance by modeling the deviation of mode sequences through temporal shifting and stretching, deriving bounds on the dual gap between transferred optimality cuts and the true optimal costs, and utilizing these bounds to quantify the level of suboptimality guaranteed in the first solve of the Benders Master Problem. Our algorithm is validated in simulation through controlling a cart-pole system with soft contact walls, a free-flying robot navigating around obstacles, and a humanoid robot standing on one leg while pushing against walls with its hands for balance. For our benchmark problems, the algorithm enumerates cuts on the order of only tens to hundreds while reaching speeds $2$-$3$ times faster than the off-the-shelf solver Gurobi, oftentimes exceeding $1000$ $Hz$. The code is available at \url{https://github.com/XuanLin/Benders-MPC}.
\end{abstract}



\begin{keyword}
Benders decomposition \sep hybrid control \sep model predictive control \sep mixed-integer quadratic programming



\end{keyword}

\end{frontmatter}



\section{Introduction}
\label{sec1}
Model Predictive Control (MPC) for hybrid systems, characterized by the presence of both continuous and discrete variables, is widely applicable to robotic motion planning and control tasks. Specific applications include control involving contact with the environment~\cite{posa2014direct, hogan2020reactive, marcucci2017approximate}, control under temporal logic constraints~\cite{lin2025towards, ren2025accelerating, kurtz2022mixed}, motion planning for robot locomotion with gait planning~\cite{aceituno2017simultaneous, zhang2021transition}, and obstacle avoidance~\cite{richards2002aircraft, marcucci2023motion}. However, the computational efficiency of hybrid MPC is often constrained due to the combinatorial complexity arising from determining the optimal discrete variable sequence, hindering its real-time robotic applications.

To address this limitation, previous works have explored several approaches. One prominent method is Explicit MPC, which entails parametrically solving the optimization problem offline to enumerate the solution space. As originally proposed by~\cite{bemporad2002explicit} for LQR problems, this technique partitions the state space into polyhedral regions where the optimal control law is piecewise affine, reducing the online computational burden. This framework was subsequently extended to hybrid systems and MIQPs in~\cite{borrelli2005dynamic, zhu2020fast}. However, due to the exponential growth in the number of polyhedral regions relative to the state dimension and prediction horizon, the application of this approach is generally limited to small-scale problems.

For fast online MPC computation, researchers have investigated smoothing methods, including approximating discrete variables with continuous variables subject to complementarity constraints~\cite{posa2014direct}, or solving for discrete variables via hierarchical optimization using smoothed gradients~\cite{le2024fast}. Similarly, smooth approximations of Signal Temporal Logic specifications have been investigated for logic-constrained motion planning~\cite{gilpin2021smooth}. These methods enable the use of efficient, off-the-shelf gradient-based nonlinear programming solvers. However, the complementarity constraints introduced to enforce integrality often violate most Constraint Qualifications established for nonlinear optimization, making robust real-time performance difficult to guarantee. Alternatively, decomposition approaches such as the Alternating Direction Method of Multipliers (ADMM)~\cite{boyd2011distributed} have been investigated to address the computational challenges of hybrid MPC~\cite{aydinoglu2023consensus, zhou2020accelerated}, where the algorithm iterates between a quadratic program and a projection step. While ADMM excels at rapidly obtaining a rough initial solution, the relaxed discrete variables may not fully converge to valid integer solutions~\cite{ren2025accelerating}. Furthermore, ADMM lacks theoretical convergence guarantees when applied to mixed-integer programming problems.

Recently, learning-based methods have emerged as a promising direction for 
accelerating hybrid MPC. Examples include learning the binary variables for 
warm start using non-parametric learning techniques~\cite{zhu2020fast} or 
fully connected neural networks~\cite{hogan2020reactive, cauligi2021coco}. 
Other research has explored training an offline neural network to warm-start 
an online primal active set solver~\cite{chen2022large}, simultaneously 
learning both the MPC policy and the dual policy that provides online 
optimality certificates~\cite{zhang2020near}, or incorporating the explicit MPC 
structure into reinforcement learning~\cite{chen2018approximating}. More 
recently, generative models have been applied to this domain, including 
diffusion models for constrained trajectory optimization~\cite{li2024constraint} 
and flow matching policies for predictive control in contact-rich 
tasks~\cite{kurtz2025generative}.

In this paper, we propose a novel hybrid MPC algorithm based on Generalized Benders decomposition (GBD) to solve control problems formulated as mixed-logical dynamical (MLD) systems. GBD separates the problem into a master problem that solves part of the variables, named complicating variables, and a subproblem that solves the rest of the variables. It uses a constraint generation technique that progressively builds representations of the feasible region and optimal cost function within the master problem through feasibility cuts and optimality cuts.

The key innovation is a warm-start technique, where the algorithm enumerates and stores cutting planes inside a finite buffer as problem instances are solved. These cuts are transferred to warm-start the next MPC iteration, avoiding the need to search from an empty set of cuts at each control cycle. On our benchmark problems, GBD often requires only a single iteration to obtain a globally optimal solution, reaching solving speeds exceeding $1000$ $Hz$. Additionally, GBD requires only tens to hundreds of accumulated cuts to provide effective warm-starts, in contrast to previous works such as \cite{cauligi2021coco} requiring over 90,000 offline training samples.

We list the contributions below:
\begin{enumerate}
    \item We propose a novel algorithm based on GBD for hybrid MPC, where cutting planes are stored inside a finite buffer and transferred across MPC iterations to provide warm-starts for new problem instances, significantly accelerating solving speeds, and \label{contri:1}
    \item We present theoretical results that bound the dual gap between transferred optimality cuts and true optimal costs, and establish conditions under which the algorithm achieves suboptimality guarantees in the first iteration, and \label{contri:2}
    \item We validate our algorithm on three robotic control scenarios: cart-pole balancing with soft contact walls, free-flying robot obstacle navigation, and a humanoid robot standing on one leg while pushing against walls with its hands for balance, demonstrating the applicability of the proposed algorithm to contact-rich tasks. \label{contri:3}
\end{enumerate}

The rest of this paper is organized as follows. Section~\ref{sec2} introduces mixed-logical dynamical systems and the GBD framework for hybrid MPC. Section~\ref{sec3} presents our warm-starting strategy through cut transfer across MPC iterations. Section~\ref{sec4} derives theoretical bounds on the dual gap between transferred optimality cuts and true optimal costs, and provides conditions that guarantee a bounded level of suboptimality in the first master problem solve. Section~\ref{sec5} validates our approach through three control scenarios: a cart-pole system with soft contact walls (Section~\ref{sec:cart_pole}), a free-flying robot navigating obstacles (Section~\ref{sec:free_flying}), and a humanoid robot balancing with wall contacts (Section~\ref{sec:humanoid_experiment}). Section~\ref{sec:conclusion} concludes with discussions and future directions.

\emph{Notations}
Vectors are bold lowercase; matrices are bold uppercase; sets are script or italicized uppercase.
The real number set is $\mathbb{R}$.
For $\boldsymbol{x}, \boldsymbol{y} \in \mathbb{R}^n$, $\boldsymbol{x} \leq \boldsymbol{y}$ indicates element-wise inequality.
For $\boldsymbol{A} \in \mathbb{R}^{n \times n}$ and $\boldsymbol{B} \in \mathbb{R}^{m \times m}$, $\text{diag}(\boldsymbol{A}, \boldsymbol{B}) \in \mathbb{R}^{(n+m) \times (n+m)}$ denotes the block diagonal matrix with diagonal blocks $\boldsymbol{A}$ and $\boldsymbol{B}$, and zeros otherwise. $\boldsymbol{I}_{n}$ denotes an identity matrix of dimension $n$. $\mathbf{1}_n$ denotes a vector of all ones of dimension $n$. Throughout the paper, we use square brackets to denote the time step for variables such as [k].


\section{Preliminaries}
\label{sec2}
\subsection{Mixed-Logical Dynamical Systems (MLDs)}

Consider mixed-logical dynamical systems that have both discrete and continuous inputs:
\begin{subequations}
\begin{align}
\boldsymbol{\dot{x}}(t) = \boldsymbol{E}_c \boldsymbol{x}(t) &+ \boldsymbol{F}_c \boldsymbol{u}(t) + \boldsymbol{G}_c \boldsymbol{\delta}(t) + \boldsymbol{n}_c(t) \\
\boldsymbol{H}_1 \boldsymbol{x}(t) &+ \boldsymbol{H}_2\boldsymbol{u}(t) + \boldsymbol{H}_3\boldsymbol{\delta}(t) \leq \boldsymbol{h} 
\end{align}
\end{subequations}
where $\mathbf{E}_c \in \mathbb{R}^{n_x \times n_x}$, $\mathbf{F}_c \in \mathbb{R}^{n_x \times n_u}$, $\mathbf{G}_c \in \mathbb{R}^{n_x \times n_\delta}$ are the system matrices, $\mathbf{H}_1 \in \mathbb{R}^{n_c \times n_x}$, $\mathbf{H}_2 \in \mathbb{R}^{n_c \times n_u}$, $\mathbf{H}_3 \in \mathbb{R}^{n_c \times n_\delta}$ are the constraint matrices, $\mathbf{h} \in \mathbb{R}^{n_c}$ is the constraint vector, and $\mathbf{n}_c(t)$ is the noise input. Here, $n_x$ is the dimension of the continuous state, $n_u$ is the dimension of the continuous input, $n_\delta$ is the dimension of the binary input, and $n_c$ is the number of inequality constraints.

We discretize this continuous system with time step $dT$, obtaining discrete dynamics with state and control constraints:
\begin{subequations} 
\begin{align}
\boldsymbol{x}[k + 1] = \boldsymbol{E} &\boldsymbol{x}[k] + \boldsymbol{F} \boldsymbol{u}[k] + \boldsymbol{G}\boldsymbol{\delta}[k] + \boldsymbol{n}[k] \label{eq:mld_dynamics} \\
\boldsymbol{H}_1 &\boldsymbol{x}[k] + \boldsymbol{H}_2\boldsymbol{u}[k] + \boldsymbol{H}_3\boldsymbol{\delta}[k] \leq \boldsymbol{h} \label{eq:mld_constraints}
\end{align}
\end{subequations}
where $\boldsymbol{x}[k] \in \mathbb{R}^{n_x}$ denotes the continuous state, $\boldsymbol{u}[k] \in \mathbb{R}^{n_u}$ denotes the continuous input, $\boldsymbol{\delta}[k] \in \{0, 1\}^{n_\delta}$ denotes the binary input, and $\boldsymbol{n}[k] \in \mathbb{R}^{n_x}$ denotes the disturbance input. The system matrices are $\boldsymbol{E} \in \mathbb{R}^{n_x \times n_x}$, $\boldsymbol{F} \in \mathbb{R}^{n_x \times n_u}$, $\boldsymbol{G} \in \mathbb{R}^{n_x \times n_\delta}$, obtained from the continuous system dynamics via standard zero-order hold discretization with sampling time $dT$.

We utilize Model Predictive Control (MPC) to solve control problems with these dynamics. MPC formulates an optimization problem under a specific initial condition $\boldsymbol{x}_0$ to compute a sequence of control inputs over a finite prediction horizon; however, it only implements the first control action. After receiving sensor feedback that provides the new initial condition, the optimization problem is solved again. The MPC formulation for the MLD system \eqref{eq:mld_dynamics}-\eqref{eq:mld_constraints} is given by:
\begin{subequations}
\begin{align}
\underset{\boldsymbol{x}[k], \boldsymbol{u}[k], \boldsymbol{\delta}[k]}{\text{minimize}} \quad & \sum_{k=0}^{N-1} \left( \|\boldsymbol{x}[k] - \boldsymbol{x}_g[k]\|^2_{\boldsymbol{Q}_k} + \|\boldsymbol{u}[k]\|^2_{\boldsymbol{R}_k} \right) + \|\boldsymbol{x}[N] - \boldsymbol{x}_g[N]\|^2_{\boldsymbol{Q}_N} \label{eq:mpc_objective}\\
\text{subject to} \quad &\boldsymbol{x}[0] = \boldsymbol{x}_0 \label{eq:mpc_initial}\\
&\boldsymbol{x}[k + 1] = \boldsymbol{E}\boldsymbol{x}[k] + \boldsymbol{F} \boldsymbol{u}[k] + \boldsymbol{G}\boldsymbol{\delta}[k] \label{eq:mpc_dynamics}\\
&\boldsymbol{H}_1\boldsymbol{x}[k] + \boldsymbol{H}_2\boldsymbol{u}[k] + \boldsymbol{H}_3\boldsymbol{\delta}[k] \leq \boldsymbol{h} \label{eq:mpc_constraints}\\
&\boldsymbol{\delta}[k] \in \{0, 1\}^{n_\delta}, \quad k = 0, \ldots, N-1 \label{eq:mpc_binary}
\end{align}
\end{subequations}
where $N$ is the prediction horizon, $\boldsymbol{x}_g[k]$ represents the reference trajectory, and $\boldsymbol{Q}_k$, $\boldsymbol{R}_k$, $\boldsymbol{Q}_N$ are positive definite weighting matrices. Equation \eqref{eq:mpc_objective}-\eqref{eq:mpc_binary} can be written in a more compact Mixed-Integer Quadratic Programming (MIQP) form:
\begin{subequations}
\begin{align}
\underset{\boldsymbol{x}, \boldsymbol{\delta}}{\text{minimize}} \quad & \|\boldsymbol{x} - \boldsymbol{x}_g\|^2_{\boldsymbol{Q}} \label{eq:compact_objective}\\
\text{subject to} \quad &\boldsymbol{A}\boldsymbol{x} = \boldsymbol{b}(\boldsymbol{x}_0, \boldsymbol{\delta}) \label{eq:compact_equality}\\
&\boldsymbol{C}\boldsymbol{x} \leq \boldsymbol{d}(\boldsymbol{\delta}) \label{eq:compact_inequality}
\end{align}
\end{subequations}
where the definitions of vectors and matrices $\boldsymbol{x}$, $\boldsymbol{\delta}$, $\boldsymbol{A}$, $\boldsymbol{b}$, $\boldsymbol{C}$, $\boldsymbol{d}$, and $\boldsymbol{Q}$ are given in \ref{app_system_matrices}.

Since MIQPs are NP-hard \cite{pia2017mixed}, real-time MPC implementation is computationally challenging. This paper applies Generalized Benders Decomposition (GBD) to solve~\eqref{eq:compact_objective}--\eqref{eq:compact_inequality}, with a novel warm-starting strategy that stores cutting planes from previous iterations to accelerate future iterations. We first review the fundamentals of GBD and its application to the MIQP formulation above.



\subsection{Generalized Benders Decomposition for MIQP}

In this paper, we apply Generalized Benders Decomposition (GBD) to solve the hybrid MPC problem \eqref{eq:compact_objective}--\eqref{eq:compact_inequality}. GBD addresses optimization problems of the following form:
\begin{subequations} \label{eq:original}
\begin{align}
    \underset{\boldsymbol{\eta}, \boldsymbol{\gamma}}{\text{minimize}} \quad & f(\boldsymbol{\eta}, \boldsymbol{\gamma}) \\
    \text{subject to} \quad & \boldsymbol{G}(\boldsymbol{\eta}, \boldsymbol{\gamma}) \leq \boldsymbol{0} \\
    & \boldsymbol{\eta} \in \mathcal{X}, \boldsymbol{\gamma} \in \mathcal{Y}
\end{align} 
\end{subequations}
where $\boldsymbol{\eta}$ denotes the complicating variables, such that fixing $\boldsymbol{\eta}$ yields a significantly easier subproblem in $\boldsymbol{\gamma}$. GBD accelerates the solution process through decomposition into two interconnected problems: a \emph{Benders Master Problem} (BMP) that proposes candidate solutions for $\boldsymbol{\eta}$, and a \emph{Benders Subproblem} (BSP) that fixes $\boldsymbol{\eta}$ and solves for the remaining variables $\boldsymbol{\gamma}$. If the BSP is infeasible, it generates a \emph{feasibility cut} that excludes the current $\boldsymbol{\eta}$ from future iterations of the BMP. If the BSP is feasible, it generates an \emph{optimality cut} that provides a lower bound on the objective function parameterized by $\boldsymbol{\eta}$. This iterative procedure continues until the gap between upper and lower bounds converges to within a specified tolerance.

For the MIQP \eqref{eq:compact_objective}--\eqref{eq:compact_inequality}, we choose the binary sequence $\boldsymbol{\delta}$ as the complicating variables following the standard GBD formulation~\cite{geoffrion1972generalized}. The BMP is:
\begin{subequations} \label{eq:bmp}
\begin{align}
    \underset{\boldsymbol{\delta}}{\text{minimize}} \quad & v(\boldsymbol{x}_0, \boldsymbol{\delta}) \label{eq:bmp_obj}\\
    \text{subject to} \quad & \boldsymbol{\delta}_k \in \{0, 1\}^{n_\delta}, \quad k = 0, \ldots, N-1 \label{eq:bmp_binary}\\
    & \boldsymbol{\delta} \in \mathcal{V}(\boldsymbol{x}_0) \label{eq:bmp_feasible}
\end{align}
\end{subequations}  
where $v(\boldsymbol{x}_0, \boldsymbol{\delta})$ denotes the optimal cost value of the BSP for a given $\boldsymbol{\delta}$, and $\mathcal{V}(\boldsymbol{x}_0)$ denotes the set of all $\boldsymbol{\delta}$ for which the BSP is feasible. The BSP fixes $\boldsymbol{\delta}$ and solves the resulting QP:
\begin{subequations} \label{eq:bsp}
\begin{align}
    v(\boldsymbol{x}_0, \boldsymbol{\delta}) = \underset{\boldsymbol{x}}{\text{minimize}} \quad & \|\boldsymbol{x} - \boldsymbol{x}_g\|^2_{\boldsymbol{Q}} \label{eq:bsp_obj}\\
    \text{subject to} \quad & \boldsymbol{A}\boldsymbol{x} = \boldsymbol{b}(\boldsymbol{x}_0, \boldsymbol{\delta}) \label{eq:bsp_eq}\\
    & \boldsymbol{C}\boldsymbol{x} \leq \boldsymbol{d}(\boldsymbol{\delta}) \label{eq:bsp_ineq}
\end{align}
\end{subequations}

The GBD algorithm iterates between the BMP and BSP. At each iteration, the BMP proposes a candidate binary sequence $\boldsymbol{\delta}_{BMP}^*$, which is then passed to the BSP. The BSP either (i) finds a feasible solution and returns an optimality cut, or (ii) proves infeasibility and returns a feasibility cut. These cuts are accumulated in the BMP to progressively refine the search space until convergence. 

Following the standard GBD framework (see, e.g., \cite{geoffrion1972generalized}), we define the cutting plane constraints:

\begin{definition} (Feasibility Cuts and Optimality Cuts). \label{def:cuts}
\begin{enumerate}
\item A \textbf{feasibility cut} $\mathcal{F} = \{\boldsymbol{\mu}^f, \boldsymbol{\pi}^f\}$ is generated from an infeasible BSP attempt under fixed $\boldsymbol{\delta}^f$, where $\boldsymbol{\mu}^f \in \mathbb{R}^{(N+1)n_x}$ and $\boldsymbol{\pi}^f \in \mathbb{R}^{Nn_c}$ are Farkas certificates corresponding to the equality and inequality constraints of the infeasible BSP, respectively. The feasibility cut takes the linear constraint form:
\begin{align}
\boldsymbol{b}(\boldsymbol{x}_0, \boldsymbol{\delta})^T \boldsymbol{\mu}^f + \boldsymbol{d}(\boldsymbol{\delta})^T \boldsymbol{\pi}^f \geq 0
\label{eq:feasibility_cut}
\end{align}
which excludes the infeasible $\boldsymbol{\delta}^f$ from the BMP.
\item An \textbf{optimality cut} $\mathcal{O} = \{(\boldsymbol{x}_0^*, \boldsymbol{\delta}^*), v^*, \boldsymbol{\mu}^*, \boldsymbol{\pi}^*\}$ is generated from a feasible BSP solution at construction point $(\boldsymbol{x}_0^*, \boldsymbol{\delta}^*)$, where $v^* = v(\boldsymbol{x}_0^*, \boldsymbol{\delta}^*)$ denotes the optimal BSP cost at this point, and $\boldsymbol{\mu}^* \in \mathbb{R}^{(N+1)n_x}$ and $\boldsymbol{\pi}^* \in \mathbb{R}^{Nn_c}$ are the dual variables corresponding to the equality and inequality constraints of the feasible BSP, respectively. The optimality cut provides a lower bound on the cost function:
\begin{align}
v(\boldsymbol{x}_0, \boldsymbol{\delta}) \geq c^* - \boldsymbol{b}(\boldsymbol{x}_0, \boldsymbol{\delta})^T \boldsymbol{\mu}^* - \boldsymbol{d}(\boldsymbol{\delta})^T \boldsymbol{\pi}^{*}
\label{eq:optimality_cut}
\end{align}
where the constant $c^* = v^* + \boldsymbol{b}(\boldsymbol{x}_0^*, \boldsymbol{\delta}^*)^T \boldsymbol{\mu}^* + \boldsymbol{d}(\boldsymbol{\delta}^*)^T \boldsymbol{\pi}^*$.
\end{enumerate}
\end{definition}

The feasibility cuts are constructed via Farkas' lemma (Theorem 4.6 in \cite{bertsimas1997introduction}). When the BSP \eqref{eq:bsp_obj}--\eqref{eq:bsp_ineq} is infeasible for some $\boldsymbol{\delta}^f$, Farkas' lemma guarantees the existence of certificates $\boldsymbol{\mu}^f \in \mathbb{R}^{(N+1)n_x}$ and $\boldsymbol{\pi}^f \in \mathbb{R}^{Nn_c}$ satisfying:
\begin{equation}
\boldsymbol{\pi}^f \geq \boldsymbol{0}, \quad \boldsymbol{A}^T \boldsymbol{\mu}^f + \boldsymbol{C}^T \boldsymbol{\pi}^f = \boldsymbol{0}, \quad \boldsymbol{b}(\boldsymbol{x}_0, \boldsymbol{\delta}^f)^T \boldsymbol{\mu}^f + \boldsymbol{d}(\boldsymbol{\delta}^f)^T \boldsymbol{\pi}^f < 0
\label{eq:farkas}
\end{equation}
The constraint~\eqref{eq:feasibility_cut} reverses this strict inequality, thereby excluding $\boldsymbol{\delta}^f$ from future BMP iterations. 

The optimality cuts are constructed via strong duality (see~\ref{app:dual_derivation} for the dual formulation of \eqref{eq:bsp_obj}--\eqref{eq:bsp_ineq}). When the BSP \eqref{eq:bsp_obj}--\eqref{eq:bsp_ineq} is feasible for some $\boldsymbol{\delta}^*$ at construction point $(\boldsymbol{x}_0^*, \boldsymbol{\delta}^*)$, the QP solver returns the optimal cost $v^* = v(\boldsymbol{x}_0^*, \boldsymbol{\delta}^*)$ and dual variables $(\boldsymbol{\mu}^*, \boldsymbol{\pi}^*)$. Strong duality at the construction point ensures $c^* = v^* + \boldsymbol{b}(\boldsymbol{x}_0^*, \boldsymbol{\delta}^*)^T \boldsymbol{\mu}^* + \boldsymbol{d}(\boldsymbol{\delta}^*)^T \boldsymbol{\pi}^*$, as \eqref{eq:optimality_cut} holds with equality at $(\boldsymbol{x}_0^*, \boldsymbol{\delta}^*)$. Since $(\boldsymbol{\mu}^*, \boldsymbol{\pi}^*)$ remains a feasible dual solution for the BSP at any $(\boldsymbol{x}_0, \boldsymbol{\delta})$, weak duality guarantees that the constraint~\eqref{eq:optimality_cut} provides a valid lower bound on $v(\boldsymbol{x}_0, \boldsymbol{\delta})$ for all feasible $\boldsymbol{\delta}$.

As the GBD algorithm iterates, it accumulates cutting planes from previous BSPs. Let $\mathcal{I}_f = \{1, 2, \ldots, |\mathcal{I}_f|\}$ denote the index set of accumulated feasibility cuts, and $\mathcal{I}_o = \{1, 2, \ldots, |\mathcal{I}_o|\}$ denote the index set of accumulated optimality cuts. The BMP progressively refines $v(\boldsymbol{x}_0, \boldsymbol{\delta})$ and $\mathcal{V}(\boldsymbol{x}_0)$ by incorporating these cuts:
\begin{subequations} \label{eq:bmp}
\begin{align}
    \underset{\boldsymbol{\delta}, z_0}{\text{minimize}} \quad & z_0 \label{eq:bmp_obj}\\
    \text{subject to} \quad & \boldsymbol{\delta}[k] \in \{0, 1\}^{n_{\delta}}, \quad k = 0, \ldots, N-1 \label{eq:bmp_binary}\\
    & \boldsymbol{b}(\boldsymbol{x}_0, \boldsymbol{\delta})^T \boldsymbol{\mu}^f_i + \boldsymbol{d}(\boldsymbol{\delta})^T \boldsymbol{\pi}^f_i \geq 0, \quad \forall i \in \mathcal{I}_f \label{eq:bmp_feas}\\
    & z_0 \geq c^*_j - \boldsymbol{b}(\boldsymbol{x}_0, \boldsymbol{\delta})^T \boldsymbol{\mu}^*_j - \boldsymbol{d}(\boldsymbol{\delta})^T \boldsymbol{\pi}^*_j, \quad \forall j \in \mathcal{I}_o \label{eq:bmp_opt}
\end{align}
\end{subequations}
where $z_0$ is an epigraph variable constrained to be above all optimality cut lower bounds. The BMP can be solved using standard Mixed-Integer Programming (MIP) solvers based on Branch-and-Bound, or via heuristic algorithms such as genetic algorithms~\cite{poojari2009improving} and greedy-backtracking methods that leverage temporal structure of BSP~\cite{lin2024accelerate}.

The GBD algorithm maintains upper and lower bounds on the optimal cost that converge as iterations proceed. The lower bound is given by the BMP objective value $LB = z_0^*$, which increases monotonically as cuts are added (assuming the BMP is solved to global optimality at each iteration). The upper bound $UB$ tracks the best feasible BSP cost found so far, which may fluctuate as the BMP proposes different binary sequences. Convergence is declared when the relative gap $g_a = |UB - LB|/|UB|$ falls below a specified tolerance $G_a$. If the BMP identifies the global optimal solution $\boldsymbol{\delta}_{BMP}^{*}$ (with respect to its lower bounds) at each iteration, this iterative refinement procedure is guaranteed to converge to the global optimum in finite iterations for MIQPs~\cite{geoffrion1972generalized}. The complete GBD algorithm without warm-starting is presented in Algorithm~\ref{alg:gbd}.

\begin{algorithm}
\caption{GBD}
\label{alg:gbd}
\KwIn{Initial condition $\boldsymbol{x}_0$, tolerance $G_a$, max iterations $I_{\max}$, initial index sets $\mathcal{I}_f = \{\}$, $\mathcal{I}_o = \{\}$}
\textbf{Initialize:} $LB \leftarrow -\infty$, $UB \leftarrow \infty$, $i \leftarrow 0$\\
\While{$|UB - LB|/|UB| \geq G_a$ and $i < I_{\max}$}{
    Solve BMP~\eqref{eq:bmp} with current $\mathcal{I}_f$, $\mathcal{I}_o$ to obtain $\boldsymbol{\delta}_{BMP}^{*}$ and $z_0^*$\\
    $LB \leftarrow z_0^*$\\
    Solve BSP~\eqref{eq:bsp} with fixed $\boldsymbol{\delta} = \boldsymbol{\delta}_{BMP}^{*}$\\
    \eIf{BSP is feasible}{
        Obtain cost $v^*$ and dual variables $(\boldsymbol{\mu}^*, \boldsymbol{\pi}^*)$, construct optimality cut using Eq.~\eqref{eq:optimality_cut}, append to $\mathcal{I}_o$\\
        \If{$v^* < UB$}{
            $UB \leftarrow v^*$, $\boldsymbol{u}^* \leftarrow \boldsymbol{u}$
        }
    }{
        Obtain Farkas certificates $(\boldsymbol{\mu}^f, \boldsymbol{\pi}^f)$, construct feasibility cut using Eq.~\eqref{eq:feasibility_cut}, append to $\mathcal{I}_f$\\
    }
    $i \leftarrow i + 1$\\
}
\Return{$\boldsymbol{u}^*$, $\mathcal{I}_f$, $\mathcal{I}_o$}
\end{algorithm}

\section{Warm-starting GBD through Cut Transfer}
\label{sec3}
In the previous section, we presented the GBD algorithm for solving a single instance of the control problem with a fixed initial condition. In this section, we consider the online MPC setting where the initial condition changes at each control iteration. As MPC proceeds online, the problem~\eqref{eq:compact_objective}--\eqref{eq:compact_inequality} needs to be constantly resolved with different $\boldsymbol{x}_0$. We show how to exploit the structure of cutting planes to warm-start subsequent problem instances, significantly accelerating the solving speeds.

The key structural observation is that $\boldsymbol{x}_0$ and $\boldsymbol{\delta}$ both enter the BSP~\eqref{eq:bsp} as parameters in the right-hand side constraint function $\boldsymbol{b}(\boldsymbol{x}_0, \boldsymbol{\delta})$, while the constraint matrices $\boldsymbol{A}$ and $\boldsymbol{C}$ remain unchanged. Consequently, both the Farkas certificates $(\boldsymbol{\mu}^f_i, \boldsymbol{\pi}^f_i)$ for feasibility cuts and the optimal dual variables $(\boldsymbol{\mu}^*_j, \boldsymbol{\pi}^*_j)$ for optimality cuts are independent of the specific value of $\boldsymbol{x}_0$. This independence is evident from examining the Farkas requirements~\eqref{eq:farkas} and the dual problem~\eqref{eq:dual_bsp}: the first two conditions of~\eqref{eq:farkas} and the inequality constraint of~\eqref{eq:dual_bsp} do not involve the parameters $(\boldsymbol{x}_0, \boldsymbol{\delta})$.

This independence enables a simple warm-starting strategy: when a new initial condition $\boldsymbol{x}_0'$ arrives, we reuse all previously enumerated cuts by updating only the parameter $\boldsymbol{x}_0$. Specifically, assume we have accumulated cutting planes with index sets $\mathcal{I}_f$ and $\mathcal{I}_o$ as in the BMP~\eqref{eq:bmp}. For the new initial condition $\boldsymbol{x}_0'$, the updated cutting plane constraints become:
\begin{align}
\boldsymbol{b}(\boldsymbol{x}_0', \boldsymbol{\delta})^T \boldsymbol{\mu}^f_i + \boldsymbol{d}(\boldsymbol{\delta})^T \boldsymbol{\pi}^f_i &\geq 0, \quad \forall i \in \mathcal{I}_f \label{eq:feasibility_cut_updated} \\
z_0 &\geq c^*_j - \boldsymbol{b}(\boldsymbol{x}_0', \boldsymbol{\delta})^T \boldsymbol{\mu}^*_j - \boldsymbol{d}(\boldsymbol{\delta})^T \boldsymbol{\pi}^*_j, \quad \forall j \in \mathcal{I}_o \label{eq:optimality_cut_updated}
\end{align}

Note that only the terms involving $\boldsymbol{b}(\cdot, \boldsymbol{\delta})$ change when the initial condition changes from $\boldsymbol{x}_0$ to $\boldsymbol{x}_0'$, while the dual certificates $\boldsymbol{\mu}^f_i, \boldsymbol{\pi}^f_i, \boldsymbol{\mu}^*_j, \boldsymbol{\pi}^*_j$ and the terms involving $\boldsymbol{d}(\boldsymbol{\delta})$ remain unchanged. By construction, the updated optimality cuts~\eqref{eq:optimality_cut_updated} provide provable lower bounds for $v(\boldsymbol{x}_0', \boldsymbol{\delta})$, and the Farkas certificates in the updated feasibility cuts~\eqref{eq:feasibility_cut_updated} continue to certify infeasibility of $\boldsymbol{\delta}$ for the BSP under $\boldsymbol{x}_0'$.

With this simple substitution, we leverage previously enumerated cuts to initialize the BMP for $\boldsymbol{x}_0'$. The algorithm then proceeds as in Algorithm~\ref{alg:gbd}, but starting with non-empty index sets $\mathcal{I}_f$ and $\mathcal{I}_o$ that provide warm-starts. Roughly speaking, when $\boldsymbol{x}_0'$ is sufficiently close to the initial conditions where cuts were originally constructed, these transferred lower bounds from the optimality cuts remain tight and the Farkas certificates remain effective at eliminating infeasible regions, thereby reducing the number of GBD iterations required to reach convergence.

To maintain computational efficiency, we limit the number of stored cuts for warm-start through a finite buffer strategy. During the solution process for a fixed $\boldsymbol{x}_0$, we allow the algorithm to enumerate as many cuts as needed to guarantee convergence to the global optimum. However, when passing cuts to subsequent $\boldsymbol{x}'_0$, we bound the storage to prevent the BMP from becoming prohibitively expensive to solve. We maintain two finite buffers with maximum capacities $K_{\text{feas}}$ and $K_{\text{opt}}$ for feasibility and optimality cuts, respectively. After each MPC iteration, newly enumerated cuts are added to the corresponding buffers. If the total number of stored cuts exceeds the buffer capacity ($|\mathcal{I}_f| > K_{\text{feas}}$ or $|\mathcal{I}_o| > K_{\text{opt}}$), the oldest cuts are removed in a first-in-first-out (FIFO) manner.

The complete MPC framework with warm-start is presented in Algorithm~\ref{alg:gbd_mpc}, and the buffer management procedure is presented in Algorithm~\ref{alg:cut_storage}.
\begin{algorithm}
\caption{GBD MPC with Warm-start}
\label{alg:gbd_mpc}
\KwIn{$K_{\text{feas}}$, $K_{\text{opt}}$, $G_a$, $I_{\max}$}
\textbf{Initialize:} $\mathcal{I}_f = \{\}$, $\mathcal{I}_o = \{\}$\\
\Loop{
    Get $\boldsymbol{x}_0$ from state estimation\\
    $\boldsymbol{u}^*$, $\mathcal{I}_{f, \text{new}}$, $\mathcal{I}_{o, \text{new}}$ $\leftarrow$ GBD($\boldsymbol{x}_0$, $G_a$, $I_{\max}$, $\mathcal{I}_f$, $\mathcal{I}_o$)\\
    Implement control $\boldsymbol{u}^*$\\
    $\mathcal{I}_f$, $\mathcal{I}_o$ $\leftarrow$ Cut\_Storage($\mathcal{I}_f$, $\mathcal{I}_o$, $\mathcal{I}_{f, \text{new}}$, $\mathcal{I}_{o, \text{new}}$, $K_{\text{feas}}$, $K_{\text{opt}}$)
}
\end{algorithm}
\begin{algorithm}
\caption{Cut Storage}
\label{alg:cut_storage}
\KwIn{$\mathcal{I}_f$, $\mathcal{I}_o$, $\mathcal{I}_{f,\text{new}}$, $\mathcal{I}_{o,\text{new}}$, $K_{\text{feas}}$, $K_{\text{opt}}$}
\ForEach{$\mathcal{F} \in \mathcal{I}_{f,\text{new}}$}
{
    Append index to $\mathcal{I}_f$ and store $\mathcal{F}$\\
    \If{$|\mathcal{I}_f| > K_{\text{feas}}$}{
        Remove oldest index from $\mathcal{I}_f$
    }
}
\ForEach{$\mathcal{O} \in \mathcal{I}_{o,\text{new}}$}
{
    Append index to $\mathcal{I}_o$ and store $\mathcal{O}$\\
    \If{$|\mathcal{I}_o| > K_{\text{opt}}$}{
        Remove oldest index from $\mathcal{I}_o$
    }
}
\Return{$\mathcal{I}_f$, $\mathcal{I}_o$}
\end{algorithm}


\section{Theoretical Bounds on the Dual Gap for Warm-Started GBD}
\label{sec4}
In this section, we derive bounds on the gap between the lower bound of the cost provided by the stored optimality cuts and the actual optimal cost under a different initial condition $\boldsymbol{x_0}'$ and the optimal binary sequence $\boldsymbol{\delta}^*$ solved by the BMP. Although $\boldsymbol{\delta}^*$ is unknown before solving the current optimization problem, we anticipate that given a limited deviation of $\boldsymbol{x}_0$ from the state $\boldsymbol{x}_{0,j}^*$ where a stored cut $j$ was constructed, the optimal $\boldsymbol{\delta}^*$ will not differ dramatically from $\boldsymbol{\delta}_j^*$. We model this similarity between $\boldsymbol{\delta}_j^*$ and $\boldsymbol{\delta}^*$ through two primary modes of variation: \emph{temporal shifting}, where contact events occur earlier or later by at most $s$ time steps, and \emph{temporal stretching}, where contact durations change by at most $r$ time steps. Additionally, we utilize Lipschitz conditions to bound cost variations with respect to changes in both $\boldsymbol{x_0}$ and $\boldsymbol{\delta}$. This approach is theoretically justified because the optimal value function of the quadratic BSP is continuous and piecewise quadratic over the bounded parameter space. Finally, we leverage these derived bounds to quantify the level of sub-optimality guaranteed in the first solve of the BMP.


\subsection{Bounding Binary Sequence Variations under Temporal Shifts and Temporal Stretches}

Consider an MPC iteration initialized at state $\boldsymbol{x}_0$. We utilize a warm-starting buffer containing a collection of stored optimality cuts, denoted by the index set $\mathcal{I}_o$. Each cut $j \in \mathcal{I}_o$ was originally constructed at a specific state $\boldsymbol{x}_{0,j}^*$ yielding an optimal binary sequence $\boldsymbol{\delta}_j^*$. We assume that with limited variation in the initial condition, $\boldsymbol{\delta}^*$ will retain similarity to the stored $\boldsymbol{\delta}_j^*$'s. Specifically, we model the deviation between a stored sequence $\boldsymbol{\delta}_j^*$ and the current optimal $\boldsymbol{\delta}^*$ through two primary types of variation:


\begin{assumption}[Temporal Shifting and Stretching]
\label{ass:temporal-modes}
The difference between any stored binary sequence $\boldsymbol{\delta}_j^*$ and the current optimal sequence $\boldsymbol{\delta}^*$ can be characterized by two types:
\begin{enumerate}
    \item \textbf{Temporal Shifting}: The sequence $\boldsymbol{\delta}^*$ differs from the original sequence $\boldsymbol{\delta}^*_j$ by a temporal shift along the time axis with maximum magnitude $s$ time steps. If shifting forward (i.e., $\boldsymbol{\delta}^*[i] = \boldsymbol{\delta}^*_j[i+s]$), identical values are placed at the end of the sequence to maintain length, and similarly for backward shifts, identical values are placed at the beginning. For example, if the original contact sequence is $[0,0,0,1,1]$, a shift by $s=2$ time steps gives $[0,1,1,1,1]$. This type captures disturbances in initial conditions that cause contact events to occur earlier or later than anticipated.
    
    \item \textbf{Temporal Stretching}: The duration of active contact periods for the stretched sequence $\boldsymbol{\delta}^*$ differs from the stored sequence $\boldsymbol{\delta}_j^*$ by at most $r$ time steps. For example, if the original contact sequence is $[0,0,1,1,0]$, a stretch by $r=2$ may give $[0,1,1,1,1]$. This type captures variations in contact duration due to factors such as different approach velocities, contact angles, or changes in the physical properties of the contacting surfaces.
    
\end{enumerate}
\end{assumption}

We introduce an additional assumption that limits the number of mode transitions within $\boldsymbol{\delta}$. This restriction is physically reasonable, as well-behaved contact strategies typically demonstrate a ``band sparse'' property: the system maintains contact for a sustained duration and then breaks contact for another sustained period, avoiding high-frequency chattering.

\begin{assumption}[Limited Mode Transitions]
\label{ass:limited-transitions}
Each binary sequence $\boldsymbol{\delta}^*$ contains at most $K$ mode transitions over the planning horizon $N$, where a transition occurs when $\boldsymbol{\delta}[k] \neq \boldsymbol{\delta}[k-1]$ for any $k$.
\end{assumption}

Assumption \ref{ass:limited-transitions} allows us to derive tighter bounds for $\|\boldsymbol{\delta}^* - \boldsymbol{\delta}^*_j\|$. Intuitively, a higher frequency of mode transitions along the time axis creates more ``edges'' where the bits in the sequence can differ from their original values under temporal shifting or stretching. With Assumptions \ref{ass:temporal-modes} and \ref{ass:limited-transitions}, we present the following result that bounds $\|\boldsymbol{\delta}^* - \boldsymbol{\delta}^*_j\|_2$ for any stored cut $j \in \mathcal{I}_o$:

\begin{lemma}[Bound on Binary Sequence Differences]
\label{lem:binary-bound}
Consider a stored binary sequence $\boldsymbol{\delta}_j^* \in \{0,1\}^{Nn_\delta}$ with at most $K$ mode transitions over the planning horizon $N$. Let $\boldsymbol{\delta}^*$ be a version of $\boldsymbol{\delta}_j^*$ that is temporally shifted by at most $s$ time steps and temporally stretched by at most $r$ total duration changes. Then the $\ell_2$-norm distance between the sequences is bounded by:
\begin{equation}
\|\boldsymbol{\delta}^* - \boldsymbol{\delta}^*_j\|_2 \leq \sqrt{(K \cdot s + r ) \cdot n_\delta}
\end{equation}
\end{lemma}
\begin{proof}
Consider a stored binary sequence $\boldsymbol{\delta}_j^*$ with $K$ mode transitions, and let $\boldsymbol{\delta}_{j,s}^*$ denote its shifted version where $\boldsymbol{\delta}_{j,s}^*[i] = \boldsymbol{\delta}_j^*[i-s]$. For a mode transition occurring at time step $k$ (where $\boldsymbol{\delta}_j^*[k-1] \neq \boldsymbol{\delta}_j^*[k]$), the shift induces discrepancies within the interval $[k, k+s-1]$. In the worst case, each time step $i \in [k, k+s-1]$ contributes $n_\delta$ to the squared $\ell_2$-norm, corresponding to a scenario where all binary variables flip (e.g., from 0 to 1). If two mode transitions are close with separation less than $s$, their difference intervals overlap, and some positions may compensate changes that reduce the total count below this upper bound. Thus, the shifting component is bounded by $K \cdot s \cdot n_\delta$.

Regarding the temporal stretching from $\boldsymbol{\delta}_{j,s}^*$ to the final sequence $\boldsymbol{\delta}^*$, this operation modifies the binary values at most $r$ time steps. Each such modification contributes at most $n_\delta$ to the squared norm. Combining these effects, the total worst-case squared difference satisfies $\|\boldsymbol{\delta}^*_j - \boldsymbol{\delta}^*\|_2^2 \leq K \cdot s \cdot n_\delta + r \cdot n_\delta$, yielding the stated bound.
\end{proof}

Equipped with this bound on the binary sequence variation, we now proceed to derive the bounds on the dual gap using Lipschitz condition.

\subsection{Lipschitz Condition of the Cost and Bounds for the Dual Gap}

In this section, we seek to bound the gap between the lower bounds provided by the stored optimality cuts and the actual optimal cost. Intuitively, this gap gauges the prediction error of the optimality cuts relative to the actual cost, which serves as a tool for guaranteeing the level of sub-optimality for BMP solutions later in the analysis. Recall that strong duality guarantees that this gap is zero at the specific initial condition and binary sequence where the cut was originally constructed. However, as the system evolves, the initial condition $\boldsymbol{x}_0$ changes, and the optimal binary solution $\boldsymbol{\delta}^*$ will also shift. Consequently, a gap emerges as a combination of how the linear cuts evolve with respect to $\boldsymbol{x}_0$ and $\boldsymbol{\delta}^*$, and how the actual cost function varies. We first utilize a Lipschitz condition to bound the variation of the cost with respect to $\boldsymbol{x}_0$ and $\boldsymbol{\delta}$:

\begin{assumption}[Lipschitz Condition]
\label{ass:lipschitz}
Assuming the parameter space of feasible $\boldsymbol{x}_0$ is bounded, the optimal cost function $v(\boldsymbol{x}_0, \boldsymbol{\delta})$ satisfies the following Lipschitz condition:
\begin{equation}
    |v(\boldsymbol{x}_0', \boldsymbol{\delta}') - v(\boldsymbol{x}_0, \boldsymbol{\delta})| \leq L_x \|\boldsymbol{x}_0' - \boldsymbol{x}_0\|_2 + L_\delta \|\boldsymbol{\delta}' - \boldsymbol{\delta}\|_2
\end{equation}
where $L_x$ and $L_\delta$ are positive Lipschitz constants.
\end{assumption}

The rationale for this assumption stems from the structure of our BSPs~\eqref{eq:bsp}, which are parametric QPs. The previous work~\cite{bemporad2002explicit} has established that the optimal value function of a multiparametric QP is continuous and piecewise quadratic. Since the binary variable $\boldsymbol{\delta}$ takes finite values, the parameter space is bounded provided that the set of feasible initial conditions $\boldsymbol{x}_0$ is bounded. Consequently, the value function has bounded gradients over this domain, thereby satisfying the Lipschitz condition.

Leveraging this Lipschitz condition, we now present a corollary that bounds the dual gap between the lower bounds from optimality cuts and the actual optimal cost under perturbed conditions:


\begin{corollary} \label{cor:dual_gap}
Consider a collection of stored optimality cuts indexed by the set $\mathcal{I}_o$ used for warm-starting. For each cut $j \in \mathcal{I}_o$, define the dual gap as $g_j(\boldsymbol{x}_0, \boldsymbol{\delta}) = v(\boldsymbol{x}_0, \boldsymbol{\delta}) - z_j(\boldsymbol{x}_0, \boldsymbol{\delta})$, where $z_j$ is the value of the optimality cut $j$ evaluated at $(\boldsymbol{x}_0, \boldsymbol{\delta})$:
\begin{equation}
z_j(\boldsymbol{x}_0, \boldsymbol{\delta}) = c^*_j - \boldsymbol{b}(\boldsymbol{x}_0, \boldsymbol{\delta})^T \boldsymbol{\mu}^*_j - \boldsymbol{d}(\boldsymbol{\delta})^T \boldsymbol{\pi}^*_j
\label{Eqn:zj_evaluation}
\end{equation}
Consider a new initial condition $\boldsymbol{x}_0$ that differs from the construction point $\boldsymbol{x}_{0,j}^*$, and any optimal binary sequence $\boldsymbol{\delta}^*$ (under $\boldsymbol{x}_0$) that differs from the stored sequence $\boldsymbol{\delta}_j^*$ by temporal shifting of at most $s$ time steps and temporal stretching of at most $r$ total duration changes. The dual gap for each cut $j \in \mathcal{I}_o$ is bounded by:
\begin{equation}
\begin{split}
0 \leq g_j(\boldsymbol{x}_0, \boldsymbol{\delta}^*) \leq & \, L_x\|\Delta \boldsymbol{x}_{0,j}\|_2 + L_{\delta} \sqrt{(K \cdot s + r )n_{\delta}} + \boldsymbol{\mu}_j^*[0]^T \Delta \boldsymbol{x}_{0,j} \\
& + \max \left( \sum_{\tau \in \mathcal{T}} \sum_{k=\tau-(s+r)}^{\tau-1} (\boldsymbol{\delta}_j^*[\tau] - \boldsymbol{\delta}_j^*[\tau-1])^T \boldsymbol{\psi}_j^*[k] \, , \, \sum_{\tau \in \mathcal{T}} \sum_{k=\tau}^{\tau+(s+r)-1} (\boldsymbol{\delta}_j^*[\tau-1] - \boldsymbol{\delta}_j^*[\tau])^T \boldsymbol{\psi}_j^*[k] \right)
\end{split}
\label{Eqn:dual_gap_bound}
\end{equation}
where $\Delta \boldsymbol{x}_{0,j} = \boldsymbol{x}_0 - \boldsymbol{x}_{0,j}^*$, $\boldsymbol{\psi}_j^*[k] = \boldsymbol{G}^T \boldsymbol{\mu}_j^*[k+1] - \boldsymbol{H}_3^T \boldsymbol{\pi}_j^*[k]$, and the set of mode transition indices $\mathcal{T} = \{ \tau \mid \boldsymbol{\delta}_j^*[\tau] \neq \boldsymbol{\delta}_j^*[\tau-1] \}$. 
\end{corollary}


\begin{proof}
Since the optimality cuts provide lower bounds for $v(\boldsymbol{x}_0, \boldsymbol{\delta})$, we have $g_j(\boldsymbol{x}_0, \boldsymbol{\delta}^*) \geq 0$. To prove the second inequality, recall that $c_j^* = v(\boldsymbol{x}_{0,j}^*, \boldsymbol{\delta}_j^*) + \boldsymbol{b}(\boldsymbol{x}_{0,j}^*, \boldsymbol{\delta}_j^*)^T \boldsymbol{\mu}_j^* + \boldsymbol{d}(\boldsymbol{\delta}_j^*)^T \boldsymbol{\pi}_j^*$. Therefore:
\begin{equation}
    z_j(\boldsymbol{x}_0, \boldsymbol{\delta}^*) = v(\boldsymbol{x}_{0,j}^*, \boldsymbol{\delta}_j^*) + [\boldsymbol{b}(\boldsymbol{x}_{0,j}^*, \boldsymbol{\delta}_j^*) - \boldsymbol{b}(\boldsymbol{x}_0, \boldsymbol{\delta}^*)]^T \boldsymbol{\mu}_j^* + [\boldsymbol{d}(\boldsymbol{\delta}_j^*) - \boldsymbol{d}(\boldsymbol{\delta}^*)]^T \boldsymbol{\pi}_j^*
\end{equation}
In addition, $v(\boldsymbol{x}_0, \boldsymbol{\delta}^*) \leq v(\boldsymbol{x}_{0,j}^*, \boldsymbol{\delta}_j^*) + L_x\|\Delta \boldsymbol{x}_{0,j}\|_2 + L_{\delta}\|\boldsymbol{\delta}^* - \boldsymbol{\delta}_j^*\|_2$ via the Lipschitz condition. Therefore:
\begin{subequations}
\label{Eqn:dual_gap_derivation}
\begin{align}
    g_j(\boldsymbol{x}_0, \boldsymbol{\delta}^*) &= v(\boldsymbol{x}_0, \boldsymbol{\delta}^*) - z_j(\boldsymbol{x}_0, \boldsymbol{\delta}^*) \nonumber \\
    &\leq L_x\|\Delta \boldsymbol{x}_{0,j}\|_2 + L_{\delta}\|\boldsymbol{\delta}^* - \boldsymbol{\delta}_j^*\|_2 - [\boldsymbol{b}(\boldsymbol{x}_{0,j}^*, \boldsymbol{\delta}_j^*) - \boldsymbol{b}(\boldsymbol{x}_0, \boldsymbol{\delta}^*)]^T \boldsymbol{\mu}_j^* - [\boldsymbol{d}(\boldsymbol{\delta}_j^*) - \boldsymbol{d}(\boldsymbol{\delta}^*)]^T \boldsymbol{\pi}_j^*\label{eq:gap_expanded} \\
    & \leq L_x\|\Delta \boldsymbol{x}_{0,j}\|_2 + L_{\delta}\|\boldsymbol{\delta}^* - \boldsymbol{\delta}_j^*\|_2 + \boldsymbol{\mu}_j^*[0]^T \Delta \boldsymbol{x}_{0,j} + \sum_{k=0}^{N-1} (\boldsymbol{\delta}^*[k] - \boldsymbol{\delta}_j^*[k])^T \boldsymbol{\psi}_j^*[k] \label{eq:gap_plugin} \\
    & \leq L_x\|\Delta \boldsymbol{x}_{0,j}\|_2 + L_{\delta} \sqrt{(K \cdot s + r )n_{\delta}} + \boldsymbol{\mu}_j^*[0]^T \Delta \boldsymbol{x}_{0,j} \nonumber \\
    & \quad + \max \left( \sum_{\tau \in \mathcal{T}} \sum_{k=\tau-(s+r)}^{\tau-1} (\boldsymbol{\delta}_j^*[\tau] - \boldsymbol{\delta}_j^*[\tau-1])^T \boldsymbol{\psi}_j^*[k] \, , \, \sum_{\tau \in \mathcal{T}} \sum_{k=\tau}^{\tau+(s+r)-1} (\boldsymbol{\delta}_j^*[\tau-1] - \boldsymbol{\delta}_j^*[\tau])^T \boldsymbol{\psi}_j^*[k] \right) \label{eq:gap_final}
\end{align}
\end{subequations}

The inequality \eqref{eq:gap_plugin} follows by substituting the definitions of $\boldsymbol{b}(\boldsymbol{x}_0, \boldsymbol{\delta})$ and $\boldsymbol{d}(\boldsymbol{\delta})$ from \ref{app_system_matrices} into the dual product terms. To reach \eqref{eq:gap_final}, we substitute the worst-case binary variation bound from Lemma \ref{lem:binary-bound}. We also rely on the fact that the difference $\boldsymbol{\delta}^*[k] - \boldsymbol{\delta}_j^*[k]$ is non-zero only in the vicinity of mode transitions. Therefore, the final summation term in \eqref{eq:gap_final} captures the worst-case inner product of these binary deviations with $\boldsymbol{\psi}_j^*[k]$, maximized over the backward and forward shifting scenarios for each transition $\tau \in \mathcal{T}$.
\end{proof}
The interpretation of the intermediate inequality \eqref{eq:gap_expanded} is straightforward. The terms $L_x\|\Delta \boldsymbol{x}_{0,j}\|_2$ and $L_{\delta}\|\boldsymbol{\delta}^* - \boldsymbol{\delta}_j^*\|_2$ bound the variation in the optimal cost function $v(\boldsymbol{x}_0, \boldsymbol{\delta})$ induced by the change in the initial condition and binary solution. The other terms correspond to the linear correction of the optimality cut along the $\boldsymbol{x}_0$ and $\boldsymbol{\delta}$ directions, scaled by the dual variables. This bound demonstrates that as the new initial condition $\boldsymbol{x}_0$ remains close to the stored construction point $\boldsymbol{x}_{0,j}^*$, the dual gap is tightly constrained, theoretically justifying the efficacy of the proposed warm-starting strategy.

\subsection{Suboptimality Bound for the First BMP Solve}

Having established the upper bound on the dual gap in the previous section, we now leverage this result to quantify the level of suboptimality guaranteed in the first solve of the BMP. The intuition is that the dual gap bounded in Corollary \ref{cor:dual_gap} represents the worst-case prediction error of the stored optimality cuts relative to the actual optimal cost. Consequently, if this prediction error is bounded within a specific tolerance, it prevents the BMP from selecting a binary sequence that is arbitrarily worse than the optimum, potentially eliminating the need for further Benders iterations. To formalize this, we first define a metric for suboptimality used in our analysis.

\begin{definition}[$\alpha$-Suboptimality] \label{def:alpha}
A feasible binary sequence $\boldsymbol{\delta}$ is said to be $\alpha$-suboptimal under $\boldsymbol{x}_0$ if the difference between its associated cost $v(\boldsymbol{x}_0, \boldsymbol{\delta})$ and the global optimal cost $v(\boldsymbol{x}_0, \boldsymbol{\delta}^*)$ satisfies:
\begin{equation}
v(\boldsymbol{x}_0, \boldsymbol{\delta}) - v(\boldsymbol{x}_0, \boldsymbol{\delta}^*) \leq \alpha
\end{equation}
where $\alpha \geq 0$ represents the level of suboptimality.
\end{definition}


With this metric defined, the following Corollary establishes a sufficient condition under which the BMP is guaranteed to return an $\alpha$-suboptimal solution in the very first iteration:




\begin{corollary}[$\alpha$-Suboptimality at First BMP Solve]
Consider the BMP solving the problem at initial condition $\boldsymbol{x}_0$, warm-started by a set of stored optimality cuts indexed by $\mathcal{I}_o$. For each cut $j \in \mathcal{I}_o$ with associated binary sequence $\boldsymbol{\delta}_j^*$, define the \textit{permissible neighborhood} $\mathcal{N}_{s,r}(\boldsymbol{\delta}_j^*)$ as the set of binary sequences obtained by applying a temporal shift of at most $s$ steps and a temporal stretch of at most $r$ duration changes to $\boldsymbol{\delta}_j^*$. Let $\mathcal{S}_{\cup} \triangleq \bigcup_{j \in \mathcal{I}_o} \mathcal{N}_{s,r}(\boldsymbol{\delta}_j^*)$ denote the union of these neighborhoods. Restricting our analysis of BMP solutions to $\mathcal{S}_{\cup}$ where warm-starting is valid (supported by Assumption~\ref{ass:temporal-modes}), it follows that the true global optimal binary sequence $\boldsymbol{\delta}^* \in \mathcal{S}_{\cup}$.

Let $\alpha \geq 0$. Assume that for any binary sequence $\boldsymbol{\delta} \in \mathcal{S}_{\cup}$, there exists at least one stored cut $j \in \mathcal{I}_o$ such that the dual gap is bounded by $\alpha$:
\begin{equation}
    g_j(\boldsymbol{x}_0, \boldsymbol{\delta}) < \alpha
\label{eq:covering_condition}
\end{equation}
where \eqref{eq:covering_condition} is verified using the bound in Corollary~\ref{cor:dual_gap}. Then, the first solve of the BMP yields a binary sequence $\boldsymbol{\delta}^*_{BMP}$ that is $\alpha$-suboptimal for the original problem, provided $\boldsymbol{\delta}^*_{BMP}$ is globally optimal with respect to the stored cuts and is feasible for the BSP.
\end{corollary}

\begin{proof}
Let $\boldsymbol{\delta}^* \in \mathcal{S}_{\cup}$ denote the true global optimal binary sequence. We need to show that for any suboptimal binary sequence $\boldsymbol{\hat\delta} \in \mathcal{S}_{\cup}$ that is strictly more than $\alpha$-suboptimal (i.e., $v(\boldsymbol{x}_0, \boldsymbol{\hat\delta}) - v(\boldsymbol{x}_0, \boldsymbol{\delta}^*) > \alpha$), the BMP will prefer $\boldsymbol{\delta}^*$ over $\boldsymbol{\hat\delta}$. Since the BMP minimizes the epigraph variable $z_0$ which is constrained by the maximum of all optimality cuts, the BMP will prefer $\boldsymbol{\delta}^*$ over $\boldsymbol{\hat\delta}$ if the cost lower bound of $\boldsymbol{\hat\delta}$ is strictly higher than that of $\boldsymbol{\delta}^*$:
\begin{equation}
    \max_{j \in \mathcal{I}_o} z_j(\boldsymbol{x}_0, \boldsymbol{\hat\delta}) > \max_{j \in \mathcal{I}_o} z_j(\boldsymbol{x}_0, \boldsymbol{\delta}^*)
\label{bmp_pf_1}
\end{equation}
Since the stored optimality cuts provide valid lower bounds on the true cost function, we have $\max_{j \in \mathcal{I}_o} z_j(\boldsymbol{x}_0, \boldsymbol{\delta}^*) \leq v(\boldsymbol{x}_0, \boldsymbol{\delta}^*)$. Therefore, a sufficient condition to guarantee \eqref{bmp_pf_1} is:
\begin{equation}
    \max_{j \in \mathcal{I}_o} z_j(\boldsymbol{x}_0, \boldsymbol{\hat\delta}) > v(\boldsymbol{x}_0, \boldsymbol{\delta}^*)
\label{bmp_pf_2}
\end{equation}

By the corollary assumption in \eqref{eq:covering_condition}, there exists a cut $\hat{j} \in \mathcal{I}_o$ such that $g_{\hat{j}}(\boldsymbol{x}_0, \hat{\boldsymbol{\delta}}) < \alpha$. Using the gap definition $z_{\hat{j}}(\boldsymbol{x}_0, \hat{\boldsymbol{\delta}}) = v(\boldsymbol{x}_0, \hat{\boldsymbol{\delta}}) - g_{\hat{j}}(\boldsymbol{x}_0, \hat{\boldsymbol{\delta}})$, we derive:
\begin{subequations}
\begin{align}
    \max_{j \in \mathcal{I}_o} z_j(\boldsymbol{x}_0, \hat{\boldsymbol{\delta}}) &\geq z_{\hat{j}}(\boldsymbol{x}_0, \hat{\boldsymbol{\delta}}) = v(\boldsymbol{x}_0, \hat{\boldsymbol{\delta}}) - g_{\hat{j}}(\boldsymbol{x}_0, \hat{\boldsymbol{\delta}}) \label{eq:z_step1} \\
    &> v(\boldsymbol{x}_0, \boldsymbol{\delta}^*) + \alpha - g_{\hat{j}}(\boldsymbol{x}_0, \hat{\boldsymbol{\delta}}) \label{eq:z_step2} \\
    &> v(\boldsymbol{x}_0, \boldsymbol{\delta}^*) \label{eq:z_step3}
\end{align}
\label{eq:z_lower_bound}
\end{subequations}
The inequality \eqref{eq:z_step2} follows from $v(\boldsymbol{x}_0, \hat{\boldsymbol{\delta}}) - v(\boldsymbol{x}_0, \boldsymbol{\delta}^*) > \alpha$, and \eqref{eq:z_step3} holds because of condition \eqref{eq:covering_condition}. We hence have shown \eqref{bmp_pf_2}, which guarantees that the BMP will prefer $\boldsymbol{\delta}^*$ over $\hat{\boldsymbol{\delta}}$. Consequently, the solution found by the BMP is guaranteed to be $\alpha$-suboptimal.
\end{proof}

This corollary provides a theoretical guarantee for the performance of the warm-started GBD. It implies that if the accumulated optimality cuts maintain a dual gap smaller than $\alpha$, the BMP is guaranteed to return an $\alpha$-suboptimal solution in the very first iteration, assuming this solution is feasible for the BSP. Provided this level of suboptimality is acceptable, this eliminates the need for further expensive Benders iterations. In the cart-pole experiment (Section \ref{sec:cart_pole}), we track the evolution of the theoretical bound derived in Corollary \ref{cor:dual_gap}. While this bound can be conservative, the warm-start strategy is highly effective empirically, where the first BMP solve frequently yields high-quality suboptimal solutions.

\section{Experiment}
\label{sec5}
We test our Benders MPC algorithm with warm-start on three different problems: controlling a cart-pole system to balance between two soft contact walls, a free-flying robot to navigate through obstacles, and a humanoid robot balancing on one leg while utilizing hand contacts with walls for stabilization. These problems are also presented as verification problems in many previous works, such as \cite{aydinoglu2023consensus, marcucci2020warm, cauligi2021coco, quirynen2023tailored, wang2018realization, marcucci2017approximate}. We implement Algorithm \ref{alg:gbd_mpc} to solve these problems. We use $G_a = 0.1$ among the proposed and benchmark methods for all problems. Other important parameters, such as $K_{feas}$ and $K_{opt}$, are chosen properly and reported for each experiment. For fair comparisons with Gurobi's MIQP solver, we use Gurobi's QP solver to solve the Benders subproblems. Note that other faster QP solvers listed by \cite{qpsolvers2023} can be implemented to further increase the solving speed. The algorithm is coded in C++ and tested on a 12th Gen Intel Core i7-12800H $\times$ 20 laptop with 16 GB of memory.

\subsection{Cart-pole with Wall Contact}
\label{sec:cart_pole}
We study the problem of controlling a cart-pole system to balance between two static soft contact walls, tested inside a PyBullet environment \cite{coumans2016pybullet}. The system dynamics, contact model, and their formulation into MLD systems follow \cite{marcucci2020warm}. The pendulum dynamics are linearized around the upright equilibrium and discretized with a step size of $dT=0.02$s. The state vector $\boldsymbol{x}[k] \in \mathbb{R}^4$ includes the cart position $x_1$, pole angle $x_2$, and their derivatives $x_3$, $x_4$. The control input $\boldsymbol{u}[k] = [f,  \lambda_1, \lambda_2 ]^T \in \mathbb{R}^3$ comprises the horizontal actuation force $f$ and contact forces $\lambda_1$, $\lambda_2$ from the right and left walls, respectively. Two binary variables $\boldsymbol{\delta}[k] \in \{0,1\}^2$ describe three contact modes: no contact, left wall contact, and right wall contact. The soft contact walls are modeled with elastic pads located at distances $d_1$ (right) and $d_2$ (left) from the origin, with stiffness $k_1$ and $k_2$. The contact logic is enforced using the standard big-M approach, as detailed in \cite{marcucci2020warm}. This problem has $n_x=4$, $n_u=3$, $n_\delta=2$, $n_c=20$. The objective function uses cost weights $\boldsymbol{Q}_k=\text{diag}(1,50,1,50)$, $\boldsymbol{R}_k=0.1\boldsymbol{I}_3$, and a terminal cost weights $\boldsymbol{Q}_N$ obtained by solving a discrete algebraic Riccati equation. The objective regulates the pole to the vertical position with zero velocities while penalizing control efforts. At the beginning of each test episode, the pendulum starts from $x_2=10^\circ$ and bumps into the wall to regain balance. Throughout each episode, persistent random disturbance torques drawn from a Gaussian distribution $\mathcal{N}(0, 8)$ $Nm$ are applied to the pole. The system must frequently contact the walls for rebalancing. All system parameters are listed in Table \ref{tab:cartpole_params}.
\begin{table}[h]
\centering
\caption{Cart-Pole System Parameters}
\label{tab:cartpole_params}
\begin{tabular}{lll}
\hline
Parameter & Symbol & Value \\
\hline
Cart mass & $m_c$ & $1.0$ $kg$ \\
Pole mass & $m_p$ & $0.4$ $kg$ \\
Pole length & $l$ & $0.6$ $m$ \\
Wall stiffness & $k_1, k_2$ & $50$ $N/m$ \\
Right wall distance & $d_1$ & $0.4$ $m$ \\
Left wall distance & $d_2$ & $0.4$ $m$ \\
Control force limit & $f_{\max}$ & $20$ $N$ \\
Angle limits & $x_2$ & $\pm\pi/2$ \\
\hline
\end{tabular}
\end{table}

\textbf{Results} 
We experimented with planning horizons $N=10$ and $N=15$, using buffer sizes $(K_{feas}, K_{opt}) = (50, 40)$ and $(150, 40)$, respectively. To evaluate performance, we ran multiple episodes and collected trajectories showing solving speed, optimal cost, GBD iterations, and the number of stored cuts. The data is collected from solved problems where at least one contact is planned. 

\begin{figure}[t!]
    \centering
    \includegraphics[width=1.0\textwidth]{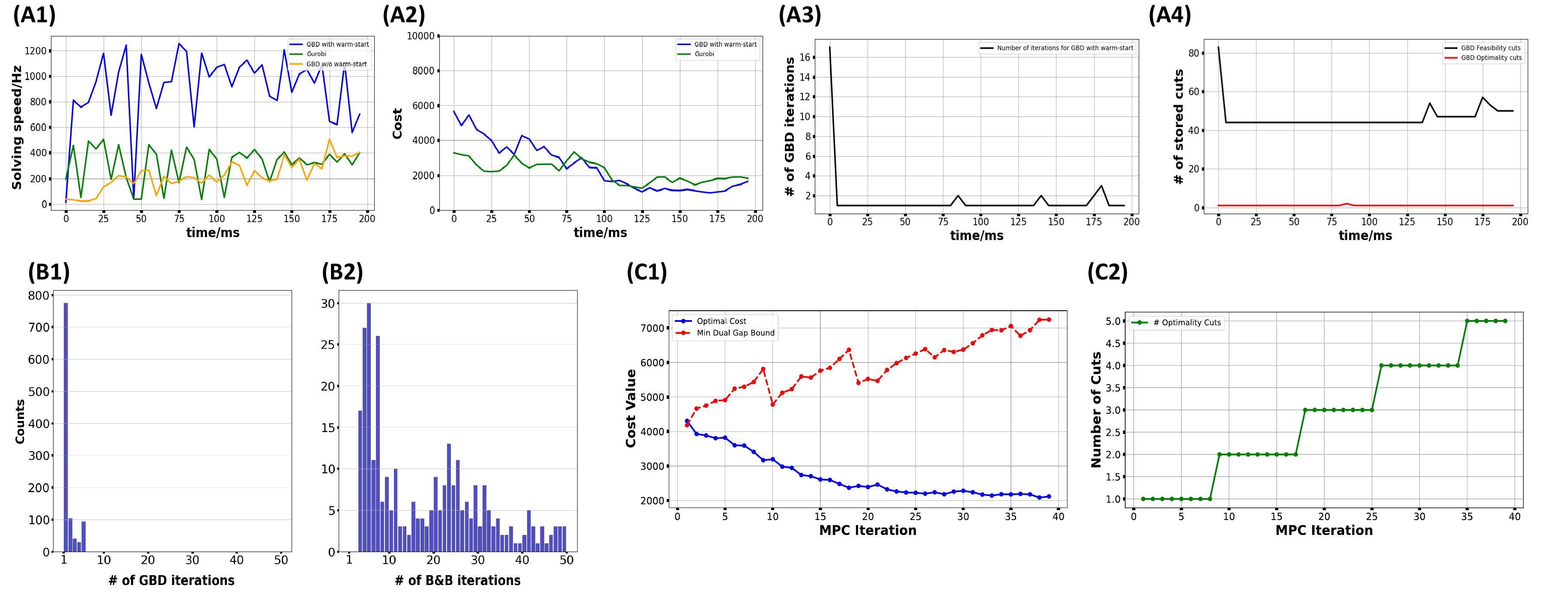}
     \caption{Cart-pole with soft walls simulated for $200$ $ms$. (A1) solving speed comparison among proposed GBD, GBD without warm-start, and Gurobi, with N=10. x-axis: simulation time in ms, y-axis: solving speed in Hz. (A2) Cost of proposed GBD and Gurobi. (A3) The number of GBD iterations. (A4) The number of cuts stored during the solving procedure. Blue curves: proposed GBD. Green curves: Gurobi. Orange curve: GBD without warm-start. Black curve: feasibility cuts, red curve: optimality cuts. (B1) Histogram result for this experiment with different $\boldsymbol{\Theta}$. x-axis: the number of GBD iterations. y axis: the count of problem instances. (B2) Same histogram result for Branch and Bound with warm-start \cite{marcucci2020warm}. (C1) The best upper bound of the dual gap among all stored optimality cuts computed from Corollary \ref{cor:dual_gap} (red curve) and global optimal cost (blue curve) at each MPC iteration, for $N=15$. (C2) The number of stored cuts for (C1) at each MPC iteration.}
     \label{fig:contact_analysis}
\end{figure}

Fig. \ref{fig:contact_analysis} shows results from a representative episode during the first $200$ $ms$ of control, where GBD begins with an empty cut set but must immediately plan contact. While GBD achieves similar cost to Gurobi (subfigure (A2)), its solving speed surpasses Gurobi after a brief cold-start phase (blue and green curves, subfigure (A1)). In contrast, the solving speed remains slower than Gurobi (orange curve, subfigure (A1)) without warm-starting. Subfigures (A3) and (A4) highlight GBD's data efficiency: fewer than 50 feasibility cuts and 5 optimality cuts suffice to provide effective warm-starts for the encountered initial conditions. After cold-start, GBD only occasionally adds new cuts, as shown by the iteration count (subfigure (A3)). This contrasts sharply with the neural-network approach of \cite{cauligi2021coco}, which requires over 90,000 offline training samples.

\begin{figure}[t!]
    \centering
    \includegraphics[width=1.0\textwidth]{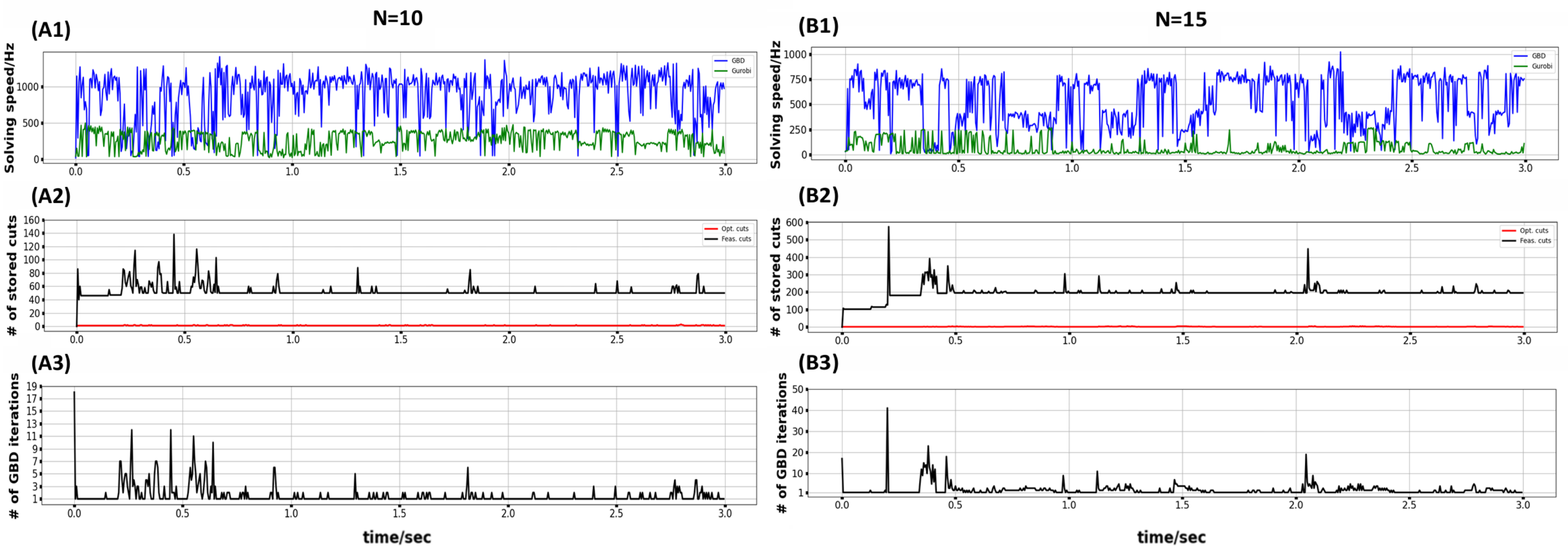}
     \caption{Solving speed, number of stored cuts, and number of GBD iterations for cart-pole with contact experiment. Left: N=10. Right: N=15. (A1), (B1) show the solving speed of GBD (blue curve) compared against Gurobi (green curve) in Hz. (A2), (B2) show the number of stored feasibility cuts (black curve) and optimality cuts (red curve). (A3), (B3) show the number of GBD iterations.}
     \label{fig:contact_long}
\end{figure}

Fig. \ref{fig:contact_long} extends this analysis to continuous control over several seconds with contact planning. The finite buffer strategy maintains solving speed by keeping the number of cuts bounded, justified by 77\% and 74\% of problems resolved in a single GBD iteration for $N=10$ and $N=15$, respectively. Throughout the horizon, GBD achieves 2-3$\times$ speedup over Gurobi on average. Computational profiling reveals that sub-QPs consume over 80\% of total solve time for both horizons, while BMPs account for less than 20\%.

Additionally, we conducted Monte Carlo analysis on 20 trajectories over the first $200$ $ms$ under random disturbance torques. The subfigures (B1) and (B2) in Fig. \ref{fig:contact_analysis} show histograms of subproblems solved at each MPC iteration for GBD and the warm-started B\&B algorithm proposed by \cite{marcucci2020warm}, respectively. Due to warm-starting, 99.2\% of GBD problem instances are solved within 5 iterations, excluding a few cold-start cases. In contrast, the B\&B algorithm requires over 10$\times$ more subproblem evaluations on average to converge, despite warm-starting reducing its subproblem count by more than 50\%.

To evaluate the quality of the bounds derived in Corollary \ref{cor:dual_gap}, we conduct an analysis for this cart-pole experiment with $N=15$. The estimation procedure consists of two phases: offline database construction and online retrieval. In the offline phase, we first sample $n=1000$ initial conditions from the state space, concentrating sampled pole positions near the contact walls. For each sampled $\boldsymbol{x}_0$, we solve the full MIQP to obtain the optimal binary sequence $\boldsymbol{\delta}^*$ and cost $v^*$. We then estimate local Lipschitz constants $L_x$ and $L_\delta$ through perturbation. For $L_x$, we solve the subproblem QP at $\boldsymbol{x}_0$ with fixed $\boldsymbol{\delta}^*$ to obtain the optimal control $\boldsymbol{u}^*[0]$, simulate one time step forward with $5$ samples of random noise to obtain perturbed states $\boldsymbol{x}_0'$, resolve the QP from each $\boldsymbol{x}_0'$ and $\boldsymbol{\delta}^*$, and compute $L_x = \max_{\boldsymbol{x}_0'} |v(\boldsymbol{x}_0', \boldsymbol{\delta}^*) - v(\boldsymbol{x}_0, \boldsymbol{\delta}^*)| / \|\boldsymbol{x}_0' - \boldsymbol{x}_0\|_2$. For $L_\delta$, we generate perturbed binary sequences by applying temporal shifts with $s=2$ and temporal stretching with $r=2$ to $\boldsymbol{\delta}^*$, solve the QP for each perturbed sequence $\boldsymbol{\delta}'$, and compute $L_\delta = \max_{\boldsymbol{\delta}'} |v(\boldsymbol{x}_0, \boldsymbol{\delta}') - v(\boldsymbol{x}_0, \boldsymbol{\delta}^*)| / \|\boldsymbol{\delta}' - \boldsymbol{\delta}^*\|_2$. The resulting database contains $(\boldsymbol{x}_0, L_x, L_\delta)$ tuples. During online MPC execution, we retrieve Lipschitz constants for a new initial condition $\boldsymbol{x}_0^{\text{query}}$ using a nearest neighbor approach based on the Euclidean distance from $\boldsymbol{x}_0^{\text{query}}$ to the stored $\boldsymbol{x}_0$ samples.

We execute the cart-pole balancing experiment for 40 MPC iterations and track the evolution of the upper bound of the dual gap $g_j(\boldsymbol{x}_0, \boldsymbol{\delta}^*)$ computed by Corollary~\ref{cor:dual_gap}.
The results are shown in subfigures (C1) and (C2) in Fig.~\ref{fig:contact_analysis}.
Subfigure (C1) plots the tightest upper bound derived from all stored optimality cuts alongside the optimal cost $v(\boldsymbol{x}_0, \boldsymbol{\delta}^*)$ throughout the trajectory, while (C2) displays the corresponding number of stored cuts.
This upper bound corresponds to the computed level of suboptimality $\alpha$ (as defined in Definition~\ref{def:alpha}), thereby quantifying the worst-case performance guarantee for the first BMP solve.
The results reveal a characteristic pattern: when a new optimality cut is enumerated, the gap bound drops significantly, then gradually increases as the initial condition $\boldsymbol{x}_0$ evolves, until the GBD algorithm enumerates a new cut that decreases the dual gap again.
Throughout the 40 MPC iterations, the theoretical dual gap bound ranges from being comparable to the cost to several times the actual optimal cost.
Despite the computed $\alpha$ often being large, the warm-start is highly effective empirically, as more than 90\% of the first feasible solutions found by the Master Problem are globally optimal.

\subsection{Free-flying Robot with Obstacle Avoidance}
\label{sec:free_flying}
\begin{figure*}[t!]
    \centering
    \includegraphics[width=0.95\textwidth]{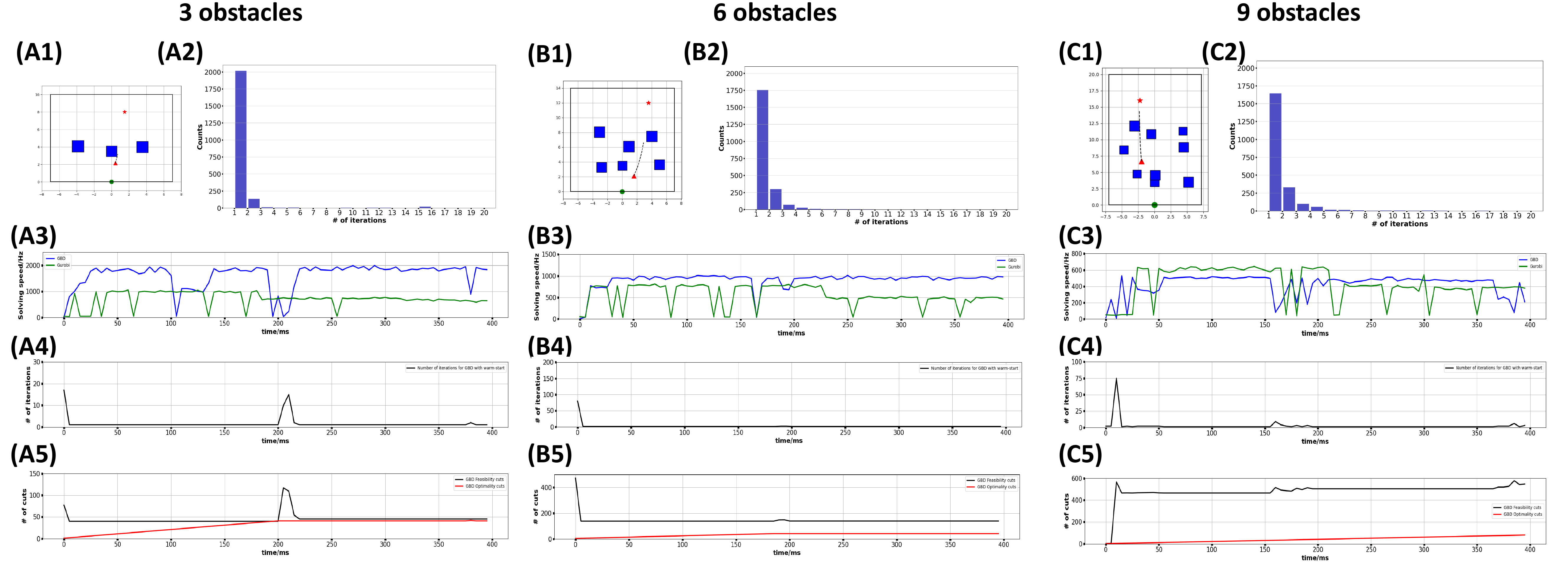}
     \caption{Results for free-flying robots navigating around 3, 6, and 9 obstacles. Sub-figure (A1), (B1), (C1) show examples of planned trajectories. The initial position is shown by green dot, target position by red star, current position by red triangle, planned trajectory in a black dashed line, and obstacles by blue squares. Sub-figure (A3)-(A5), (B3)-(B5), (C3)-(C5) show solving speed, number of GBD iterations, and number of stored cuts, corresponding to (A1), (B1), (C1). Sub-figure (A2), (B2), (C2) show histograms for the number of GBD iterations generated by Monte Carlo experiments.}
     \label{fig:free_flying_all}
\end{figure*}

We study a free-flying robot navigating through 2-D obstacles to reach a goal position under linear point-mass dynamics. Three example scenarios are shown in subfigures (A1), (B1), (C1) of Fig.~\ref{fig:free_flying_all}. The state vector $\boldsymbol{x}[k] \in \mathbb{R}^4$ consists of the 2-D position $(x, y)$ and velocity $(\dot{x}, \dot{y})$. The control input $\boldsymbol{u}[k] \in \mathbb{R}^2$ consists of pushing forces in the $x$ and $y$ directions. For each obstacle, we define two binary variables $\boldsymbol{\delta}_i[k] \in \{0,1\}^2$ to encode which side the robot passes. Each square obstacle divides the free space around it into four regions: $\mathcal{R}_{e,1}$ (right), $\mathcal{R}_{e,2}$ (top), $\mathcal{R}_{e,3}$ (left), and $\mathcal{R}_{e,4}$ (bottom). The binary encoding $\boldsymbol{\delta}_i[k] \in \{00, 01, 10, 11\}$ enforces that the robot position $(x[k], y[k])$ lies in the corresponding region $\mathcal{R}_{e,1}, \mathcal{R}_{e,2}, \mathcal{R}_{e,3}, \mathcal{R}_{e,4}$, respectively. These logical constraints are formulated as mixed-integer linear inequalities using the standard big-M method. The complete constraint set includes obstacle avoidance, state bounds, and control limits. For $M_o$ obstacles, the problem dimensions are $n_x=4$, $n_u=2$, $n_{\delta}=2M_o$, and $n_c=4M_o+8$. Here, $4M_o$ constraints enforce obstacle avoidance (four linear inequalities per obstacle defining the regions $\mathcal{R}_{e,j}$), 4 constraints enforce box bounds on velocities $(v_{\min} \leq \dot{x} \leq v_{\max}, v_{\min} \leq \dot{y} \leq v_{\max})$, and 4 constraints enforce bilateral limits on control forces $(u_{\min} \leq u_x \leq u_{\max}, u_{\min} \leq u_y \leq u_{\max})$. The objective function minimizes a weighted sum of the Euclidean distance to the goal and control effort.

The task of our hybrid MPC is to generate control inputs that guide the robot to a target position while avoiding obstacles under disturbance forces. We select a robot mass of 1 kg with control force limits of $\pm 30$ N in both $x$ and $y$ directions. The system dynamics are discretized with time step $dT=0.02$ s. To create diverse test scenarios, obstacles are initially placed on a uniform grid and then randomly perturbed to avoid clustering. The obstacle width $d_o$ is sampled from a Gaussian distribution $d_o \sim \mathcal{N}(0.7, 0.05)$ meters. The target position has a uniformly sampled $x$-coordinate, while the $y$-coordinate is set beyond all obstacles to require navigation through the obstacle field. Persistent disturbance forces are applied throughout each trajectory, with magnitudes sampled from $\mu \sim \mathcal{N}(0, 10)$ $N$ independently in both $x$ and $y$ directions. For the objective function, we choose state weights $\boldsymbol{Q}_k=\text{diag}(100,100,1,1)$, and control weights $\boldsymbol{R}_k=\boldsymbol{I}_2$. The terminal cost matrix $\boldsymbol{Q}_N$ is computed by solving the discrete-time algebraic Riccati equation.

\textbf{Results}
We evaluate performance across increasing problem scales by varying the number of obstacles $M_o \in \{3, 6, 9, 12\}$ with corresponding planning horizons $N \in \{9, 12, 15, 18\}$. Buffer sizes are chosen as $(K_{\text{feas}}, K_{\text{opt}}) = (50, 50)$ for $N=9$, $(150, 50)$ for $N=12$, and $(500, 50)$ for $N=15, 18$. Results are compiled in Fig.~\ref{fig:free_flying_all}.

Subfigures (A1), (B1), and (C1) show representative planned trajectories (dotted lines) with 3, 6, and 9 obstacles. Subfigures (A3)-(A5), (B3)-(B5), and (C3)-(C5) plot solving speed, number of GBD iterations, and number of stored cuts over time for these trajectories. To assess statistical performance, we conducted Monte Carlo experiments with 20 trajectories under randomized obstacle positions, target locations, and disturbances. The iteration count histograms are shown in subfigures (A2), (B2), and (C2). For 3 and 6 obstacles, GBD achieves faster average solving speeds than Gurobi. For 9 and 12 obstacles, the average solving speeds are comparable. Subfigures (A5), (B5), and (C5) highlight the data efficiency of our approach: fewer than 200 stored cuts provide effective warm-starts across the encountered initial conditions. This contrasts with \cite{cauligi2021coco}, which requires over 90,000 offline training samples to achieve similar warm-starting capabilities.

\subsection{Humanoid Balancing on One Leg with Wall Contact}
\label{sec:humanoid_experiment}
In this section, we evaluate the proposed algorithm on a realistic scenario where a humanoid robot stands on one leg while utilizing bilateral wall contacts for stabilization. This scenario represents a practical application of hybrid dynamics for humanoids to plan the timing and forces for making or breaking hand contact.

\begin{figure*}[t!]
    \centering
    \includegraphics[width=0.9\textwidth]{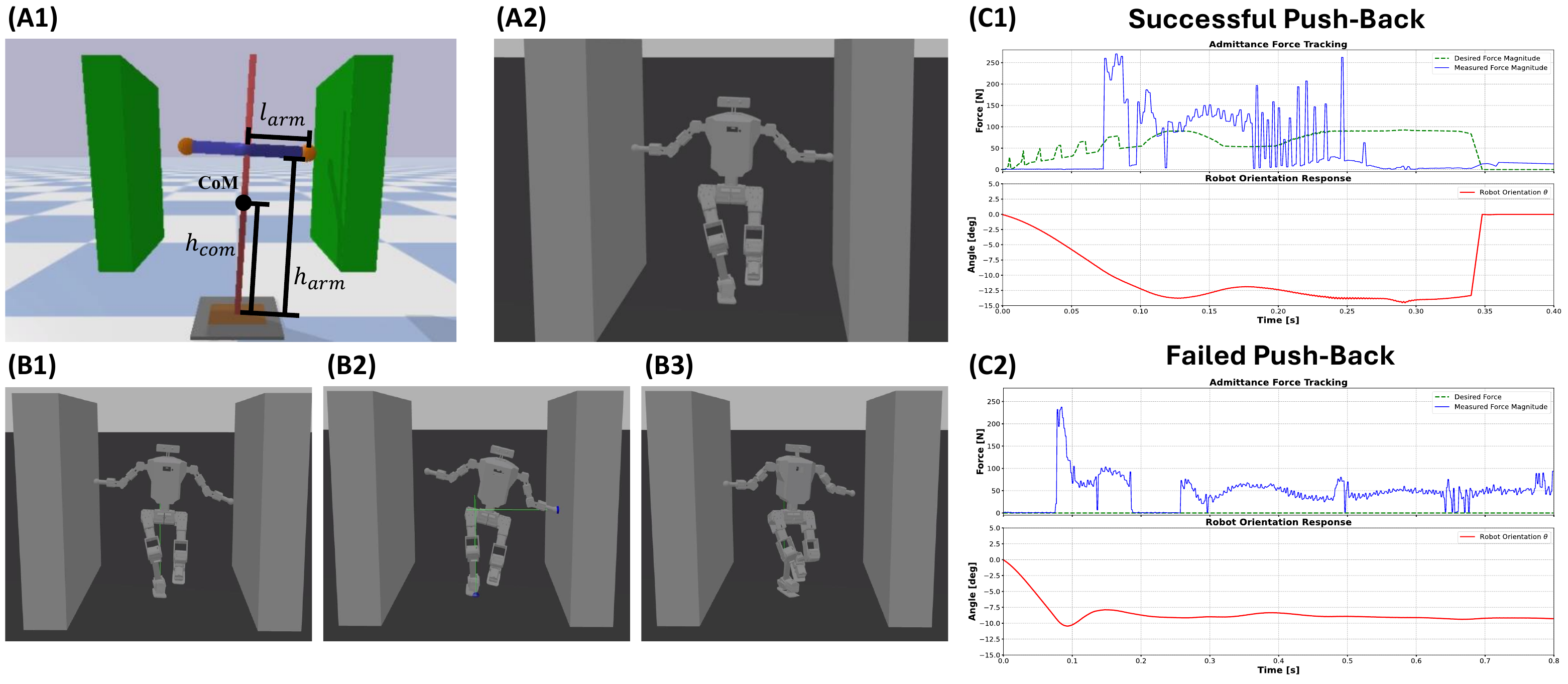}
     \caption{Humanoid balancing with wall contacts. (A1) Simplified inverted pendulum model with bilateral walls for MPC planning. (A2) Full humanoid robot with bilateral walls in Gazebo simulation. (B1-B3) Consecutive snapshots of disturbance recovery: the robot leans right, pushes off the wall to generate restoring moment, and returns to balance. The blue dot in (B2) shows the contact point and the green lines indicate the contact forces. (C1) Admittance force tracking performance (top) and robot lean angle trajectory (bottom) for a successful push-back recovery, where the measured forces (blue curve) track the planned forces (green curve) from the simplified model. (C2) A comparison case using pure position control for the arm, where the robot fails to generate the necessary push-off force and remains stuck leaning against the wall.}
     \label{fig:humanoid_setup}
\end{figure*}

To enable real-time MPC, we use a simplified pendulum model with contact for planning. The system is modeled as an inverted pendulum with an actuated pivot representing the ankle joint, as illustrated in Fig.~\ref{fig:humanoid_setup}(A1). A rigid link of mass $m$ represents the robot body, with the center of mass (CoM) located at distance $h_{\text{com}}$ from the pivot. A horizontal bar of length $2l_{\text{arm}}$ with two contact points at height $h_{\text{arm}}$ represents the arms that can push against walls located at distances $d_R$ (right) and $d_L$ (left) from the vertical centerline where the stance foot is located. Since $l_{\text{arm}} < |d_R|=|d_L|$, contact occurs only when the body tilts sufficiently toward a wall, requiring the controller to plan when to make or break contact and determine the contact forces. The state vector is $\boldsymbol{x} = [\theta, \dot{\theta}]^T$, where $\theta$ is the lean angle from vertical. The control inputs are $\boldsymbol{u} = [\tau_a, f_R, f_L]^T$, where $\tau_a$ is the ankle torque and $f_R$, $f_L$ are the contact forces from the right and left walls, respectively. The model parameters, including mass $m$, centroidal inertia $I_{\text{com}}$, and maximum ankle torque $\tau_{\max}$, are selected to approximate the physical characteristics of the HiTorque MiniHi humanoid robot. All system parameters are listed in Table~\ref{tab:params}.

The MPC problem on the simplified pendulum model generates desired ankle torques and wall contact forces in real-time. To validate the control on a realistic platform, we implement the controller in Gazebo simulation with a full humanoid model (HiTorque MiniHi, as shown in Fig.~\ref{fig:humanoid_setup}(A2)). A QP-based whole-body controller~\eqref{eqn:wbc} provides basic stability control and tracks the desired ankle torque commands from the MPC. The desired wall contact forces are tracked by admittance controllers on the robot arms~\eqref{eqn:admittance}. Similar to the cart-pole experiment, persistent random disturbance torques are applied to both the simplified model and the full humanoid throughout the simulation to challenge the controller's robustness.
\begin{table}[h]
\centering
\caption{System Parameters for Humanoid Balancing}
\label{tab:params}
\begin{tabular}{lccc}
\toprule
\textbf{Parameter} & \textbf{Symbol} & \textbf{Value} & \textbf{Unit} \\
\midrule
Mass & $m$ & 25.0 & kg \\
CoM Height & $h_{com}$ & 0.4 & m \\
Arm Contact Height & $h_{arm}$ & 0.6 & m \\
Arm Length & $l_{arm}$ & 0.2 & m \\
Centroidal Inertia & $I_{com}$ & 0.8 & kg$\cdot$m$^2$ \\
Max Ankle Torque & $\tau_{max}$ & 7.0 & Nm \\
Max Contact Force & $F_{max}$ & 200.0 & N \\
Right Wall Dist. & $d_R$ & 0.5 & m \\
Left Wall Dist. & $d_L$ & -0.5 & m \\
\bottomrule
\end{tabular}
\end{table}

We derive the MLD formulation for the simplified inverted pendulum model with contacts described above. The equation of motion governing the rotational dynamics about the pivot is linearized around the upright equilibrium $\theta = 0$, yielding:
\begin{equation}
    I_{com} \ddot{\theta} = m g h_{com} \theta + \tau_{a} + h_{arm}(f_{L} - f_{R})
    \label{eq:pendulum_dynamics_linearized}
\end{equation} 

The control problem is subject to physical limitations on the ankle actuator, ground friction, and the logic of wall contacts. The horizontal ground reaction force $f_{f}$ is determined by the linear momentum balance of the CoM horizontal acceleration ($\ddot{x}_{com} \approx h_{com}\ddot{\theta}$):
\begin{equation}
    f_f + f_L - f_R = m h_{com} \ddot{\theta}
    \label{eq:momentum_balance}
\end{equation}
Substituting $\ddot{\theta}$ from Eq.~\eqref{eq:pendulum_dynamics_linearized} allows us to express $f_f$ as a linear function of states and inputs. We enforce the Coulomb friction limit assuming a constant vertical reaction force $N \approx mg$ and coefficient of friction $\mu$, resulting in two inequality constraints:
\begin{subequations}
\begin{align}
    \frac{m^2 g h_{com}^2}{I_{com}} \theta + \frac{m h_{com}}{I_{com}} \tau_{a} - \left(1 - \frac{m h_{com} h_{arm}}{I_{com}}\right) f_L + \left(1 - \frac{m h_{com} h_{arm}}{I_{com}}\right) f_R &\le \mu mg \\
    -\frac{m^2 g h_{com}^2}{I_{com}} \theta - \frac{m h_{com}}{I_{com}} \tau_{a} + \left(1 - \frac{m h_{com} h_{arm}}{I_{com}}\right) f_L - \left(1 - \frac{m h_{com} h_{arm}}{I_{com}}\right) f_R &\le \mu mg
\end{align}
\label{eq:friction_constraints}
\end{subequations}


The contact forces $f_R, f_L$ are decision variables governed by the contact logic. We introduce binary variables $\delta_{R}[k], \delta_{L}[k] \in \{0,1\}$ to indicate whether the right or left wall is contacted, respectively. If contact occurs ($\delta_{R}[k]=1$ or $\delta_{L}[k]=1$), the corresponding contact force is allowed to be non-zero; otherwise, the force must be zero. Additionally, the geometric constraint ensures that the binary variable activates only when the arm reaches the wall. Using the Big-M formulation, these conditions are expressed as:
\begin{subequations}
\begin{align}
0 \le f_{R}[k] &\le F_{max} \delta_{R}[k] \label{eq:contact_force_R}\\
0 \le f_{L}[k] &\le F_{max} \delta_{L}[k] \label{eq:contact_force_L}\\
h_{arm}\theta[k] &\ge (d_R-l_{arm}) - M_g(1 - \delta_{R}[k]) \label{eq:contact_geo_R}\\
h_{arm}\theta[k] &\le (d_L+l_{arm}) + M_g(1 - \delta_{L}[k]) \label{eq:contact_geo_L}
\end{align}
\label{eq:contact_constraints}
\end{subequations}
where $F_{max}$ is the wall contact force limit, and $M_g$ is a sufficiently large constant for the Big-M formulation of the geometric constraints.

The complete MLD formulation includes the linearized dynamics \eqref{eq:pendulum_dynamics_linearized}, friction constraints \eqref{eq:friction_constraints}, contact logic \eqref{eq:contact_constraints}, and torque limits $|\tau_a[k]| \le \tau_{max}$. The full discrete-time system matrices are given in~\ref{app:humanoid_mld}. The optimization objective minimizes a quadratic cost $J = \sum_k ||\boldsymbol{x}[k]||^2_{\boldsymbol{Q}} + ||\boldsymbol{u}[k]||^2_{\boldsymbol{R}}$ that regulates the pendulum to $\theta=0$ with zero angular velocity, similar to Section~\ref{sec:cart_pole}.


To maintain robot balance and track the desired contact torque from the simplified pendulum model, the whole-body controller computes joint accelerations $\boldsymbol{\ddot{q}}$, joint torques $\boldsymbol{\tau}$, and contact forces $\boldsymbol{f}_j$ that are dynamically consistent with the full robot model. Let $n_a$ be the number of actuated joints and $n = n_a + 6$, where the additional 6 degrees of freedom represent the floating base. The equations of motion for the humanoid robot are~\cite{featherstone2008rigid}:
\begin{equation}
\boldsymbol{M}\boldsymbol{\ddot{q}} + \boldsymbol{C}\boldsymbol{\dot{q}} + \boldsymbol{G} = \boldsymbol{S}^T \boldsymbol{\tau} + \sum_{j=1}^{N_c} \boldsymbol{J}_{c_j}^T \boldsymbol{f}_j
\end{equation}
where $\boldsymbol{q} \in \mathbb{R}^{n}$ is the vector of generalized coordinates, $\boldsymbol{M} \in \mathbb{R}^{n \times n}$ is the inertia matrix, $\boldsymbol{C} \in \mathbb{R}^{n}$ is the vector of centrifugal and Coriolis terms, $\boldsymbol{G} \in \mathbb{R}^{n}$ is the gravity vector, $\boldsymbol{S} \in \mathbb{R}^{n \times n_a}$ is the actuation selection matrix, $\boldsymbol{\tau} \in \mathbb{R}^{n_a}$ is the joint torque vector, $\boldsymbol{J}_{c_j} \in \mathbb{R}^{3 \times n}$ and $\boldsymbol{f}_j \in \mathbb{R}^{3}$ are respectively the contact Jacobian and contact force at the $j$th contact point. Since this problem is planar and we want to track the ankle contact torque $\tau_a$ from the simplified model, we model the foot with two contact points at the sides of the foot plate that generate vertical contact forces $\boldsymbol{f}_j$. The resultant moment from these two forces about the ankle joint produces the desired torque. Hence, the number of contact points is $N_c=2$.

Given desired ankle torque $\tau_a$ from the simplified pendulum model and setting desired base acceleration $\boldsymbol{\ddot{x}}_{base}=\boldsymbol{J}_{base} \boldsymbol{\ddot{q}} + \boldsymbol{\dot{J}}_{base} \boldsymbol{\dot{q}}=\boldsymbol{0}$ to maintain stability, the whole-body controller solves the following weighted QP:
\begin{equation}
\begin{aligned}
\underset{\boldsymbol{\ddot{q}}, \boldsymbol{\tau}, \boldsymbol{f}_j}{\text{minimize}} \quad & \left\| \boldsymbol{J}_{base} \boldsymbol{\ddot{q}} + \boldsymbol{\dot{J}}_{base} \boldsymbol{\dot{q}} \right\|^2_{\boldsymbol{W}_{base}} + \left\| \boldsymbol{r}^T \boldsymbol{f} - \tau_a \right\|^2_{W_{\tau}} \\
\text{subject to} \quad & \boldsymbol{S}_f \left( \boldsymbol{M}\boldsymbol{\ddot{q}} + \boldsymbol{C}\boldsymbol{\dot{q}} + \boldsymbol{G} - \boldsymbol{S}^T \boldsymbol{\tau} - \sum_{j=1}^{N_c} \boldsymbol{J}_{c_j}^T \boldsymbol{f}_j \right) = \boldsymbol{0} \\
& \boldsymbol{\tau}_{min} \leq \boldsymbol{\tau} \leq \boldsymbol{\tau}_{max} \\
& \boldsymbol{f}_j \in \mathcal{C}_j, \quad j = 1, \ldots, N_c
\end{aligned}
\label{eqn:wbc}
\end{equation}
where $\boldsymbol{J}_{base}$ is the base Jacobian, $\boldsymbol{r} = [d, -d]^T$ is the moment arm vector relating contact forces $\boldsymbol{f} = [f_1, f_2]^T$ to ankle torque, $\boldsymbol{W}_{base}$, $W_{\tau}$ are weight matrices, $\boldsymbol{S}_f$ is the floating base selection matrix that enforces only the floating base dynamics as equality constraints, and $\mathcal{C}_j$ represents the friction cone constraint for contact force $\boldsymbol{f}_j$. The computed joint accelerations $\boldsymbol{\ddot{q}}$ and joint torques $\boldsymbol{\tau}$ are then sent to low-level motor controllers for tracking.


To track the desired contact forces on the hands, we utilize admittance control on both arms. The admittance controller measures the actual contact force $\boldsymbol{f}_{meas}$ via wrist force-torque sensors in simulation, and controls the arm position to achieve the reference contact forces $\boldsymbol{F}_{ref}$ ($f_L$ or $f_R$) generated by the simplified pendulum model:
\begin{equation}
\boldsymbol{M}_d\boldsymbol{\ddot{x}} + \boldsymbol{D}_d\boldsymbol{\dot{x}} = \boldsymbol{K}_f(\boldsymbol{f}_{meas} - \boldsymbol{F}_{ref})
\label{eqn:admittance}
\end{equation}
where $\boldsymbol{M}_d$, $\boldsymbol{D}_d$, and $\boldsymbol{K}_f$ are the desired mass, damping, and force gain matrices, respectively. Integrating the resulting acceleration yields velocity commands $\boldsymbol{\dot{x}}$, which are mapped to joint velocities via the manipulator Jacobian, allowing the arms to track the planned contact forces on the wall.

\textbf{Results} We experimented with planning horizon $N=10$ and $(K_{feas}, K_{opt}) = (50, 40)$. Random disturbance torques uniformly distributed in $[-10, 10]$ $Nm$ were continuously applied to the pendulum throughout the simulation. Without disturbance torques, the robot steadily maintains one-foot balance in the upright position. Under random disturbances, the robot constantly tilts toward one side, triggering the MPC to plan contact forces that allow the robot to push against the wall and regain balance. Similar to the cart-pole experiment, GBD's solving speed consistently exceeds $1000$ $Hz$ enabling real-time planning of contact forces. Fig. \ref{fig:humanoid_setup}(B1-B3) shows three consecutive moments of this recovery behavior, where the robot leans toward the right wall, makes contact, pushes off to generate a restoring moment, and returns toward the vertical equilibrium. Fig. \ref{fig:humanoid_setup}(C1) and (C2) present a comparison experiment to validate the proposed control strategy. (C1) illustrates the tracking performance of the desired wall contact forces by the admittance controller alongside the robot's lean angle, demonstrating that the admittance control effectively generates the necessary restoring moment to return the robot to the upright position. In (C2), we disable the admittance controller such that the arm relies solely on position control to hold its pose. As a result, when the robot leans to one side, the arms fail to generate sufficient push-off force to restore equilibrium. The robot becomes stuck leaning against the wall, where the recorded contact force corresponds merely to the passive reaction force generated by the robot's leaning weight.

These results underscore the practical utilization of the proposed GBD algorithm. The hybrid MPC based on GBD is computationally lightweight and capable of fast re-planning. Despite its simplicity, the inverted pendulum with contact serves as an effective model of the humanoid's dominant dynamics during wall-supported balancing. This allows the high-level planner to provide physically consistent target forces in real-time for the low-level tracking controller to achieve push recovery on the full humanoid robot.


\section{Conclusion, Discussion, and Future Work}
\label{sec:conclusion}
In this paper, we present a novel Hybrid MPC algorithm based on Generalized Benders Decomposition to efficiently solve MIQP control problems with contact constraints. The core innovation of our approach is a warm-starting strategy that accumulates feasibility and optimality cuts in a finite buffer and transfers them across MPC iterations, avoiding the computational burden of resolving problems from scratch. We provide a theoretical analysis of this warm-starting performance by modeling mode sequence deviations through temporal shifting and stretching. This analysis allows us to derive bounds on the dual gap and quantify the level of suboptimality guaranteed during the first solve of the Benders Master Problem. We validated the proposed algorithm through three distinct robotic scenarios: a cart-pole system contacting soft walls, a free-flying robot navigating around obstacles, and a humanoid robot maintaining balance on a single leg via wall contacts. The experimental results demonstrate that our method requires only tens to hundreds of stored cuts to generate effective warm-starts, in contrast to learning-based methods that may require thousands of training samples. Furthermore, the algorithm consistently attained high solving speeds, frequently running 2-3 times faster than the commercial solver Gurobi.

One limitation of our current theoretical analysis is the assumption that the bounds for temporal shifting and stretching, denoted by $s$ and $r$, are predetermined constants. In practice, the necessary magnitude of these deviations is linked to the variation in the initial condition, $\boldsymbol{x}_0$, between iterations. This variation, in turn, depends on the intensity of process noise and the MPC update frequency, which dictates how far the system state evolves during the computation window. Future investigations could explore probabilistic models or learning-based classifiers to predict the likelihood of mode transitions based on $\boldsymbol{x}_0$, establishing an explicit functional relationship of $(s,r)$.

We are currently working to deploy the controller for humanoid balancing on a single leg with wall contact on the physical HiTorque MiniHi hardware. While simulation results confirm the controller's capability, bridging the sim-to-real gap requires fine-tuning the low-level whole-body controllers and force controllers to ensure the hardware can accurately track the contact forces demanded by the MPC.

Finally, the application of Generalized Benders Decomposition for Hybrid MPC is not limited to robotic control with contacts. Control problems involving discrete logical mode changes, such as gait selection, or those governed by temporal logic constraints, align naturally with the MIQP structure addressed by our framework. Furthermore, the underlying concept of accumulating constraints and lower bounds offers a powerful mechanism for general online learning. Future research may explore extending this idea to learn safety constraints within uncertain dynamic environments, where the optimization formulation cannot be pre-designed and requires real-time adaptation.

\bibliographystyle{elsarticle-num} 
\bibliography{reference}

@article{geoffrion1972generalized,
  title={Generalized benders decomposition},
  author={Geoffrion, Arthur M},
  journal={Journal of optimization theory and applications},
  volume={10},
  pages={237--260},
  year={1972},
  publisher={Springer}
}

@article{marcucci2020warm,
  title={Warm start of mixed-integer programs for model predictive control of hybrid systems},
  author={Marcucci, Tobia and Tedrake, Russ},
  journal={IEEE Transactions on Automatic Control},
  volume={66},
  number={6},
  pages={2433--2448},
  year={2020},
  publisher={IEEE}
}

@book{bertsimas1997introduction,
  title={Introduction to linear optimization},
  author={Bertsimas, Dimitris and Tsitsiklis, John N},
  volume={6},
  year={1997},
  publisher={Athena scientific Belmont, MA}
}

@article{aydinoglu2023consensus,
  title={Consensus Complementarity Control for Multi-Contact MPC},
  author={Aydinoglu, Alp and Wei, Adam and Posa, Michael},
  journal={arXiv preprint arXiv:2304.11259},
  year={2023}
}

@article{bemporad2002explicit,
  title={The explicit linear quadratic regulator for constrained systems},
  author={Bemporad, Alberto and Morari, Manfred and Dua, Vivek and Pistikopoulos, Efstratios N},
  journal={Automatica},
  volume={38},
  number={1},
  pages={3--20},
  year={2002},
  publisher={Elsevier}
}

@article{zhu2020fast,
  title={Fast non-parametric learning to accelerate mixed-integer programming for hybrid model predictive control},
  author={Zhu, Jia-Jie and Martius, Georg},
  journal={IFAC-PapersOnLine},
  volume={53},
  number={2},
  pages={5239--5245},
  year={2020},
  publisher={Elsevier}
}

@inproceedings{marcucci2017approximate,
  title={Approximate hybrid model predictive control for multi-contact push recovery in complex environments},
  author={Marcucci, Tobia and Deits, Robin and Gabiccini, Marco and Bicchi, Antonio and Tedrake, Russ},
  booktitle={2017 IEEE-RAS 17th International Conference on Humanoid Robotics (Humanoids)},
  pages={31--38},
  year={2017},
  organization={IEEE}
}

@article{cauligi2021coco,
  title={Coco: Online mixed-integer control via supervised learning},
  author={Cauligi, Abhishek and Culbertson, Preston and Schmerling, Edward and Schwager, Mac and Stellato, Bartolomeo and Pavone, Marco},
  journal={IEEE Robotics and Automation Letters},
  volume={7},
  number={2},
  pages={1447--1454},
  year={2021},
  publisher={IEEE}
}

@article{hogan2020reactive,
  title={Reactive planar non-prehensile manipulation with hybrid model predictive control},
  author={Hogan, Francois R and Rodriguez, Alberto},
  journal={The International Journal of Robotics Research},
  volume={39},
  number={7},
  pages={755--773},
  year={2020},
  publisher={SAGE Publications Sage UK: London, England}
}

@article{coumans2016pybullet,
  title={Pybullet, a python module for physics simulation for games, robotics and machine learning},
  author={Coumans, Erwin and Bai, Yunfei},
  year={2016}
}

@inproceedings{zhang2021transition,
  title={Transition motion planning for multi-limbed vertical climbing robots using complementarity constraints},
  author={Zhang, Jingwen and Lin, Xuan and Hong, Dennis W},
  booktitle={2021 IEEE International Conference on Robotics and Automation (ICRA)},
  pages={2033--2039},
  year={2021},
  organization={IEEE}
}

@article{marcucci2023motion,
  title={Motion planning around obstacles with convex optimization},
  author={Marcucci, Tobia and Petersen, Mark and von Wrangel, David and Tedrake, Russ},
  journal={Science robotics},
  volume={8},
  number={84},
  pages={eadf7843},
  year={2023},
  publisher={American Association for the Advancement of Science}
}

@article{quirynen2023tailored,
  title={Tailored presolve techniques in branch-and-bound method for fast mixed-integer optimal control applications},
  author={Quirynen, Rien and Di Cairano, Stefano},
  journal={Optimal Control Applications and Methods},
  volume={44},
  number={6},
  pages={3139--3167},
  year={2023},
  publisher={Wiley Online Library}
}

@software{qpsolvers2023,
  author = {Caron, Stéphane and Arnström, Daniel and Bonagiri, Suraj and Dechaume, Antoine and Flowers, Nikolai and Heins, Adam and Ishikawa, Takuma and Kenefake, Dustin and Mazzamuto, Giacomo and Meoli, Donato and O'Donoghue, Brendan and Oppenheimer, Adam A. and Pandala, Abhishek and Quiroz Omaña, Juan José and Rontsis, Nikitas and Shah, Paarth and St-Jean, Samuel and Vitucci, Nicola and Wolfers, Soeren and @bdelhaisse and @MeindertHH and @rimaddo and @urob and @shaoanlu},
  license = {LGPL-3.0},
  month = dec,
  title = {{qpsolvers: Quadratic Programming Solvers in Python}},
  url = {https://github.com/qpsolvers/qpsolvers},
  version = {4.2.0},
  year = {2023}
}

@inproceedings{wang2018realization,
  title={Realization of a real-time optimal control strategy to stabilize a falling humanoid robot with hand contact},
  author={Wang, Shihao and Hauser, Kris},
  booktitle={2018 IEEE International Conference on Robotics and Automation (ICRA)},
  pages={3092--3098},
  year={2018},
  organization={IEEE}
}

@book{featherstone2008rigid,
  title={Rigid Body Dynamics Algorithms},
  author={Featherstone, Roy},
  year={2008},
  publisher={Springer}
}

@article{pia2017mixed,
  title={Mixed-integer quadratic programming is in NP},
  author={Pia, Alberto Del and Dey, Santanu S and Molinaro, Marco},
  journal={Mathematical Programming},
  volume={162},
  number={1},
  pages={225--240},
  year={2017},
  publisher={Springer}
}

@article{posa2014direct,
  title={A direct method for trajectory optimization of rigid bodies through contact},
  author={Posa, Michael and Cantu, Cecilia and Tedrake, Russ},
  journal={The International Journal of Robotics Research},
  volume={33},
  number={1},
  pages={69--81},
  year={2014},
  publisher={Sage Publications Sage UK: London, England}
}

@article{lin2025towards,
  title={Towards Tighter Convex Relaxation of Mixed-integer Programs: Leveraging Logic Network Flow for Task and Motion Planning},
  author={Lin, Xuan and Ren, Jiming and Luo, Yandong and Xie, Weijun and Zhao, Ye},
  journal={arXiv preprint arXiv:2509.24235},
  year={2025}
}

@article{ren2025accelerating,
  title={Accelerating signal-temporal-logic-based task and motion planning of bipedal navigation using benders decomposition},
  author={Ren, Jiming and Lin, Xuan and Mineyev, Roman and Feigh, Karen M and Coogan, Samuel and Zhao, Ye},
  journal={arXiv preprint arXiv:2508.13407},
  year={2025}
}

@article{kurtz2022mixed,
  title={Mixed-integer programming for signal temporal logic with fewer binary variables},
  author={Kurtz, Vincent and Lin, Hai},
  journal={IEEE Control Systems Letters},
  volume={6},
  pages={2635--2640},
  year={2022},
  publisher={IEEE}
}

@article{aceituno2017simultaneous,
  title={Simultaneous contact, gait, and motion planning for robust multilegged locomotion via mixed-integer convex optimization},
  author={Aceituno-Cabezas, Bernardo and Mastalli, Carlos and Dai, Hongkai and Focchi, Michele and Radulescu, Andreea and Caldwell, Darwin G and Cappelletto, Jos{\'e} and Grieco, Juan C and Fern{\'a}ndez-L{\'o}pez, Gerardo and Semini, Claudio},
  journal={IEEE Robotics and Automation Letters},
  volume={3},
  number={3},
  pages={2531--2538},
  year={2017},
  publisher={IEEE}
}

@inproceedings{richards2002aircraft,
  title={Aircraft trajectory planning with collision avoidance using mixed integer linear programming},
  author={Richards, Arthur and How, Jonathan P},
  booktitle={Proceedings of the 2002 American control conference (IEEE Cat. No. CH37301)},
  volume={3},
  pages={1936--1941},
  year={2002},
  organization={IEEE}
}

@article{borrelli2005dynamic,
  title={Dynamic programming for constrained optimal control of discrete-time linear hybrid systems},
  author={Borrelli, Francesco and Baoti{\'c}, Mato and Bemporad, Alberto and Morari, Manfred},
  journal={Automatica},
  volume={41},
  number={1},
  pages={17--33},
  year={2005},
  publisher={Elsevier}
}

@article{gilpin2021smooth,
  title={A Smooth Robustness Measure of Signal Temporal Logic for Symbolic Control},
  author={Gilpin, Yann and Kurtz, Vince and Lin, Hai},
  journal={IEEE Control Systems Letters},
  volume={5},
  number={1},
  pages={241--246},
  year={2021},
  publisher={IEEE}
}

@article{le2024fast,
  title={Fast contact-implicit model predictive control},
  author={Le Cleac'h, Simon and Howell, Taylor A and Yang, Shuo and Lee, Chi-Yen and Zhang, John and Bishop, Arun and Schwager, Mac and Manchester, Zachary},
  journal={IEEE Transactions on Robotics},
  volume={40},
  pages={1617--1629},
  year={2024},
  publisher={IEEE}
}

@article{boyd2011distributed,
  title={Distributed optimization and statistical learning via the alternating direction method of multipliers},
  author={Boyd, Stephen and Parikh, Neal and Chu, Eric and Peleato, Borja and Eckstein, Jonathan and others},
  journal={Foundations and Trends{\textregistered} in Machine learning},
  volume={3},
  number={1},
  pages={1--122},
  year={2011},
  publisher={Now Publishers, Inc.}
}

@inproceedings{zhou2020accelerated,
  title={Accelerated admm based trajectory optimization for legged locomotion with coupled rigid body dynamics},
  author={Zhou, Ziyi and Zhao, Ye},
  booktitle={2020 American Control Conference (ACC)},
  pages={5082--5089},
  year={2020},
  organization={IEEE}
}

@inproceedings{li2024constraint,
  title={Constraint-aware diffusion models for trajectory optimization},
  author={Li, Anjian and Ding, Zihan and Dieng, Adji Bousso and Beeson, Ryne},
  booktitle={International Conference on Dynamic Data Driven Applications Systems},
  pages={308--316},
  year={2024},
  organization={Springer}
}

@article{kurtz2025generative,
  title={Generative predictive control: Flow matching policies for dynamic and difficult-to-demonstrate tasks},
  author={Kurtz, Vince and Burdick, Joel W},
  journal={arXiv preprint arXiv:2502.13406},
  year={2025}
}

@article{zhang2020near,
  title={Near-optimal rapid MPC using neural networks: A primal-dual policy learning framework},
  author={Zhang, Xiaojing and Bujarbaruah, Monimoy and Borrelli, Francesco},
  journal={IEEE Transactions on Control Systems Technology},
  volume={29},
  number={5},
  pages={2102--2114},
  year={2020},
  publisher={IEEE}
}

@inproceedings{chen2018approximating,
  title={Approximating explicit model predictive control using constrained neural networks},
  author={Chen, Steven and Saulnier, Kelsey and Atanasov, Nikolay and Lee, Daniel D and Kumar, Vijay and Pappas, George J and Morari, Manfred},
  booktitle={2018 Annual American control conference (ACC)},
  pages={1520--1527},
  year={2018},
  organization={IEEE}
}

@article{chen2022large,
  title={Large scale model predictive control with neural networks and primal active sets},
  author={Chen, Steven W and Wang, Tianyu and Atanasov, Nikolay and Kumar, Vijay and Morari, Manfred},
  journal={Automatica},
  volume={135},
  pages={109947},
  year={2022},
  publisher={Elsevier}
}

@article{poojari2009improving,
  title={Improving benders decomposition using a genetic algorithm},
  author={Poojari, Chandra A and Beasley, John E},
  journal={European Journal of Operational Research},
  volume={199},
  number={1},
  pages={89--97},
  year={2009},
  publisher={Elsevier}
}

@article{lin2024accelerate,
  title={Accelerate hybrid model predictive control using generalized benders decomposition},
  author={Lin, Xuan},
  journal={arXiv preprint arXiv:2406.00780v1},
  year={2024},
  note={Version 1. Available at \url{https://arxiv.org/abs/2406.00780v1}}
}

\appendix
\section{Vector and Matrix Definitions for Compact MIQP Formulation}
\label{app_system_matrices}
The compact form \eqref{eq:compact_objective}-\eqref{eq:compact_inequality} is obtained by stacking variables and constraints over the prediction horizon. Define the concatenated variables:
\begin{align}
\boldsymbol{x} &= \begin{bmatrix} \boldsymbol{x}[0]^T & \boldsymbol{u}[0]^T & \cdots & \boldsymbol{x}[N-1]^T & \boldsymbol{u}[N-1]^T & \boldsymbol{x}[N]^T \end{bmatrix}^T \in \mathbb{R}^{N(n_x+n_u)+n_x}\\
\boldsymbol{\delta} &= \begin{bmatrix} \boldsymbol{\delta}[0]^T & \boldsymbol{\delta}[1]^T & \cdots & \boldsymbol{\delta}[N-1]^T \end{bmatrix}^T \in \mathbb{R}^{Nn_\delta}
\end{align}

The dynamics constraint matrix is:
\begin{align}
\boldsymbol{A} = \begin{bmatrix}
\boldsymbol{I}_{n_x} & \boldsymbol{0} & & & \\
-\boldsymbol{E} & -\boldsymbol{F} & \boldsymbol{I}_{n_x} & \boldsymbol{0} & \\
& & -\boldsymbol{E} & -\boldsymbol{F} & \ddots \\
& & & \ddots & \ddots & \boldsymbol{I}_{n_x} & \boldsymbol{0} \\
& & & & -\boldsymbol{E} & -\boldsymbol{F} & \boldsymbol{I}_{n_x}
\end{bmatrix} \in \mathbb{R}^{(N+1)n_x \times (N(n_x+n_u)+n_x)}
\end{align}

The right-hand side vector is:
\begin{align}
\boldsymbol{b}(\boldsymbol{x}_0, \boldsymbol{\delta}) = \begin{bmatrix} \boldsymbol{x}_0^T & (\boldsymbol{G}\boldsymbol{\delta}[0])^T & \cdots & (\boldsymbol{G}\boldsymbol{\delta}[N-1])^T \end{bmatrix}^T \in \mathbb{R}^{(N+1)n_x}
\end{align}

The inequality constraint matrix is:
\begin{align}
\boldsymbol{C} = \begin{bmatrix}
\boldsymbol{H}_1 & \boldsymbol{H}_2 & & & \\
& & \boldsymbol{H}_1 & \boldsymbol{H}_2 & \\
& & & \ddots & \ddots \\
& & & & \boldsymbol{H}_1 & \boldsymbol{H}_2 & \boldsymbol{0}
\end{bmatrix} \in \mathbb{R}^{Nn_c \times (N(n_x+n_u)+n_x)}
\end{align}

The inequality constraint right-hand side is:
\begin{align}
\boldsymbol{d}(\boldsymbol{\delta}) = \begin{bmatrix} (\boldsymbol{h} - \boldsymbol{H}_3\boldsymbol{\delta}[0])^T & \cdots & (\boldsymbol{h} - \boldsymbol{H}_3\boldsymbol{\delta}[N-1])^T \end{bmatrix}^T \in \mathbb{R}^{Nn_c}
\end{align}

The objective weight matrix is:
\begin{align}
\boldsymbol{Q} = \text{diag}(\boldsymbol{Q}_0, \boldsymbol{R}_0, \ldots, \boldsymbol{Q}_{N-1}, \boldsymbol{R}_{N-1}, \boldsymbol{Q}_N) \in \mathbb{R}^{(N(n_x+n_u)+n_x) \times (N(n_x+n_u)+n_x)}
\end{align}

\section{Dual Formulation of the Benders Subproblem}
\label{app:dual_derivation}
We present the dual problem of the BSP~\eqref{eq:bsp} to establish the foundation for optimality cuts. Introducing dual variables $\boldsymbol{\mu} \in \mathbb{R}^{(N+1)n_x}$ for the equality constraints and $\boldsymbol{\pi} \in \mathbb{R}^{Nn_c}$ for the inequality constraints, the dual problem is:
\begin{equation}
\begin{aligned}
d(\boldsymbol{x}_0, \boldsymbol{\delta}) = \underset{\boldsymbol{\mu}, \boldsymbol{\pi}}{\text{maximize}} \quad & -\frac{1}{4}\|\boldsymbol{A}^T\boldsymbol{\mu} + \boldsymbol{C}^T\boldsymbol{\pi}\|^2_{\boldsymbol{Q}^{-1}} + \boldsymbol{x}_g^T(\boldsymbol{A}^T\boldsymbol{\mu} + \boldsymbol{C}^T\boldsymbol{\pi}) - \boldsymbol{b}(\boldsymbol{x}_0, \boldsymbol{\delta})^T\boldsymbol{\mu} - \boldsymbol{d}(\boldsymbol{\delta})^T\boldsymbol{\pi} \\
\text{subject to} \quad & \boldsymbol{\pi} \geq \boldsymbol{0}
\end{aligned}
\label{eq:dual_bsp}
\end{equation}
Weak duality guarantees $d(\boldsymbol{x}_0, \boldsymbol{\delta}) \leq v(\boldsymbol{x}_0, \boldsymbol{\delta})$ for any $(\boldsymbol{x}_0, \boldsymbol{\delta})$. Since the BSP is a convex QP with linear constraints, strong duality holds (e.g. under Slater's condition), giving $d(\boldsymbol{x}_0, \boldsymbol{\delta}) = v(\boldsymbol{x}_0, \boldsymbol{\delta})$ at optimality.


\section{MLD Matrices for Humanoid Balancing}
\label{app:humanoid_mld}
The complete MLD formulation for the simplified inverted pendulum model with contacts used by the humanoid balancing experiment is given below, where the state is $\boldsymbol{x} = [\theta, \dot{\theta}]^T$, the control input is $\boldsymbol{u} = [\tau_a, f_R, f_L]^T$, and the binary variables are $\boldsymbol{\delta} = [\delta_R, \delta_L]^T$:

\begin{equation}
E = \begin{bmatrix} 1 & dT \\ \frac{mgh_{com}dT}{I_{com}} & 1 \end{bmatrix}, \quad
F = \begin{bmatrix} 0 & 0 & 0 \\ \frac{dT}{I_{com}} & -\frac{h_{arm}dT}{I_{com}} & \frac{h_{arm}dT}{I_{com}} \end{bmatrix}, \quad
G = \boldsymbol{0}
\end{equation}

\begin{equation}
H_1 = \begin{bmatrix} 
\frac{m^2gh^2_{com}}{I_{com}} & 0 \\ 
-\frac{m^2gh^2_{com}}{I_{com}} & 0 \\ 
0 & 0 \\ 
0 & 0 \\ 
0 & 0 \\ 
0 & 0 \\ 
0 & 0 \\ 
0 & 0 \\
-h_{arm} & 0 \\ 
h_{arm} & 0 
\end{bmatrix} \quad
H_2 = \begin{bmatrix} 
\frac{mh_{com}}{I_{com}} & 1-\frac{mh_{com}h_{arm}}{I_{com}} & \frac{mh_{com}h_{arm}}{I_{com}}-1 \\ 
-\frac{mh_{com}}{I_{com}} & \frac{mh_{com}h_{arm}}{I_{com}}-1 & 1-\frac{mh_{com}h_{arm}}{I_{com}} \\ 
1 & 0 & 0 \\ 
-1 & 0 & 0 \\ 
0 & -1 & 0 \\ 
0 & 0 & -1 \\ 
0 & 1 & 0 \\ 
0 & 0 & 1 \\ 
0 & 0 & 0 \\ 
0 & 0 & 0 
\end{bmatrix} \quad
H_3 = \begin{bmatrix} 
0 & 0 \\ 
0 & 0 \\ 
0 & 0 \\ 
0 & 0 \\ 
0 & 0 \\ 
0 & 0 \\ 
-F_{max} & 0 \\ 
0 & -F_{max} \\ 
M_g & 0 \\ 
0 & M_g 
\end{bmatrix} \quad
h = \begin{bmatrix} 
\mu mg \\ 
\mu mg \\ 
\tau_{max} \\ 
\tau_{max} \\ 
0 \\ 
0 \\ 
0 \\ 
0 \\ 
-(d_R - l_{arm}) + M_g \\ 
(d_L + l_{arm}) + M_g 
\end{bmatrix}
\end{equation}



\end{document}